\documentclass[runningheads]{llncs}
\usepackage[T1]{fontenc}
\usepackage{graphicx}
\usepackage{hyperref}
\hypersetup{
    colorlinks=true,
    linkcolor=blue,
    filecolor=magenta,      
    urlcolor=cyan,
    pdftitle={Overleaf Example},
    pdfpagemode=FullScreen,
    }

\usepackage{listings}
\usepackage{amsmath,amssymb,amsfonts}
\usepackage{textcomp}
\usepackage[utf8]{inputenc}
\usepackage{microtype}
\usepackage[a-2b]{pdfx}
\usepackage{graphicx}
\usepackage{balance}  
\usepackage[tight,footnotesize]{subfigure}
\usepackage{multirow}
\usepackage{algorithm}
\usepackage{algorithmic}
\usepackage{listings}
\usepackage{wrapfig}
\usepackage{float}
\usepackage{xcolor}
\usepackage{outlines}
\usepackage[normalem]{ulem}
\usepackage{cleveref}
\usepackage{enumitem}
\usepackage{soul}
\usepackage{cuted, lipsum}
\usepackage[listings,skins,breakable]{tcolorbox}
\usepackage{booktabs}
\usepackage{tabularx}
\usepackage{subcaption}

\newcommand{\eat}[1]{}
\usepackage[symbol]{footmisc}

\usepackage{color}
\begin{document}

\title{IDNet: A Novel Dataset for Identity Document Analysis and Fraud Detection  }

%
\titlerunning{IDNet}

\author{Hong Guan*\inst{1} \and
Yancheng Wang*\inst{1} \and
Lulu Xie* \inst{1} \and
Soham Nag*\inst{1} \and
Rajeev Goel \inst{1} \and
Niranjan Erappa Narayana Swamy \inst{1} \and
Yingzhen Yang \inst{1} \and \\
Chaowei Xiao \inst{2} \and
Jonathan Prisby \inst{3} \and
Ross Maciejewski \inst{1} \and
Jia Zou \inst{1}}

\authorrunning{Guan et al.}
%
\institute{Arizona State University \and
University of Wisconsin--Madison \and
Department of Homeland Security Science and Technology Directorate}


%
\maketitle              
\def\thefootnote{*}\footnotetext{These authors contributed equally to this work}\def\thefootnote{\arabic{footnote}}

\begin{abstract}
Effective fraud detection and analysis of government-issued identity documents, such as passports, driver's licenses, and identity cards, are essential in thwarting identity theft and bolstering security on online platforms. The training of accurate fraud detection and analysis tools depends on the availability of extensive identity document datasets. However, current publicly available benchmark datasets for identity document analysis, including MIDV-500, MIDV-2020, and FMIDV, fall short in several aspects: they offer a limited number of samples, cover insufficient varieties of fraud patterns, and seldom include alterations in critical personal identifying fields like portrait images, limiting their utility in training models capable of detecting realistic frauds while preserving privacy.
In response to these shortcomings, our research introduces a new benchmark dataset, IDNet, designed to advance privacy-preserving fraud detection efforts. The IDNet dataset comprises \textcolor{black}{$837,060$} images of synthetically generated identity documents, totaling approximately \textcolor{black}{$490$} gigabytes, categorized into $20$ types from $10$ U.S. states and $10$ European countries. We evaluate the utility and present use cases of the dataset, illustrating how it can aid in training privacy-preserving fraud detection methods, facilitating the generation of camera and video capturing of identity documents, and testing schema unification and other identity document management functionalities.

\end{abstract}

\section{Introduction}

The surge in digital platforms offering remote identity {proofing} has escalated concerns regarding the forgery of identity documents, including passports, driver's licenses, and identity cards. The Financial Crimes Enforcement Network (FinCEN) reported that in 2021, around $1.6$ million Bank Secrecy Act (BSA) reports—constituting $42\%$ of all reports filed that year—were related to identity fraud, highlighting \$$212$ billion in suspicious transactions~\cite{FinCEN}. This issue poses risks across various sectors, including finance, healthcare, travel, retail, government, telecommunications, and gambling~\cite{onfido}. According to a recent industry analysis~\cite{onfido}, fraudulent techniques have evolved from simple forgeries, such as name alterations, to the advanced use of generative AI/ML technologies for creating deceptive images, like face morphing~\cite{venkatesh2021face}. Notably, most remote identity {validation} services on these digital platforms rely on images captured in white light rather than multi-spectral imaging techniques like near-infrared and ultra-violet light. This paper examines the efficacy of identity {validation} under these common lighting conditions, which is the focus scenario of this paper.

Despite the availability of several public datasets for identity document analysis, which focused primarily on images taken in white light, such as MIDV-500~\cite{arlazarov2019midv}, MIDV-2019~\cite{bulatov2020midv}, MIDV-2020~\cite{bulatovich2022midv}, FMIDV~\cite{al2023guilloche}, BID~\cite{de2020bid}, and  SIDTD~\cite{boned2024synthetic}, our examination has uncovered significant limitations in these resources.

\noindent
$\bullet$ \textbf{Limited number of distinct samples:} Most existing datasets contain less than $1,000$ distinct identity documents. While these datasets may help develop tools for simple tasks such as optical character recognition (OCR), they are insufficient for training and testing AI/ML models for complicated tasks such as privacy-preserving analysis. Although the BID dataset contains $28,800$ distinct identity documents, the portrait photos are blurred. This characteristic makes it unsuitable for critical tasks such as detecting face morphing and portrait substitution, where clear images are essential for accurate model performance.


\noindent
$\bullet$ \textbf{Insufficient Fraud Patterns}: Only a {few} publicly available datasets, FMIDV~\cite{al2023guilloche} and SIDTD~\cite{boned2024synthetic}, which build upon the MIDV dataset, contain identity documents with fraudulent alterations. FMIDV presents a sole Copy-and-Move fraud pattern, where guilloche patterns are replicated and repositioned among documents. Conversely, SIDTD employs basic Crop-and-Move with inpainting techniques to simulate fraudulent activity. Nevertheless, fraud techniques such as face morphing, portrait substitution, and the intricate alteration of textual data remain unrepresented in these public collections. Crucially, as privacy issues take center stage in identity document management, the introduction of complex fraud patterns that intersect with extensive personal identifier information (PII), like portrait photos, ghost images, dates of birth, names, and addresses, is imperative for honing privacy-centric fraud detection methodologies. If fraud patterns were not intruding upon PII fields, redacting these fields in existing publicly available benchmarks could help preserve privacy during model training. The creation and availability of a new benchmark dataset containing representative fraud patterns are pivotal for enhancing the precision and confidentiality aspects of fraud detection in complex scenarios.

These constraints significantly hamper the progression of cutting-edge techniques for identity document analysis and fraud detection in an era increasingly dominated by AI/ML methodologies~\cite{kamuangu2024review}~\cite{wang2021research}~\cite{alghofaili2020financial} and where privacy considerations are paramount in the application of these technologies~\cite{fritz2022financial}~\cite{bernabe2020aries}. Overcoming these limitations is challenging because of the substantial costs of creating synthetic datasets that accurately mimic a wide range of real-life identity documents, which amounts to counterfeit documents. This difficulty is further compounded by the fact that one of the core objectives of organizations responsible for designing and issuing identity documents is to combat counterfeiting.

 \begin{figure}[!htbp]
\begin{center}
  \includegraphics[width=\linewidth]{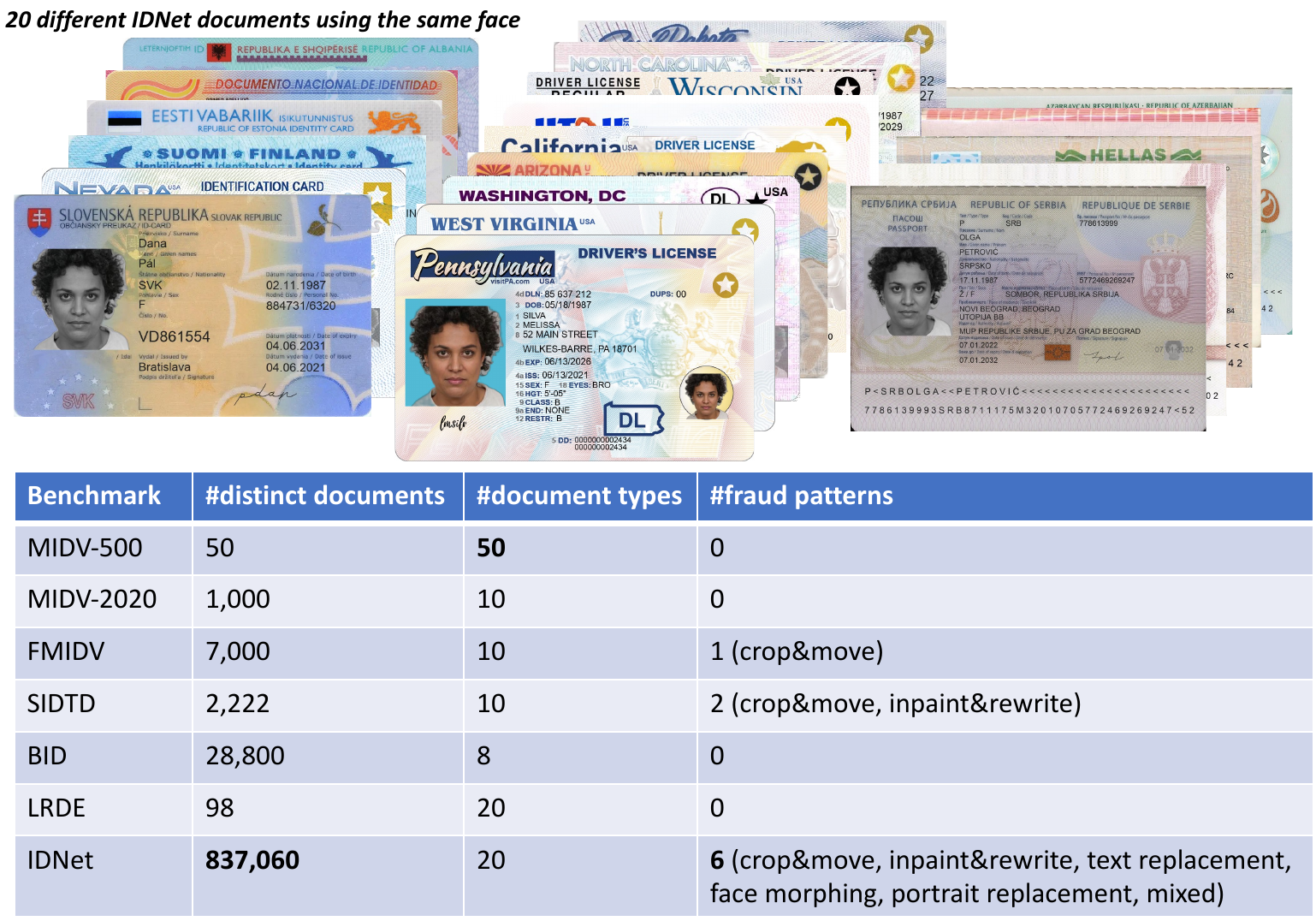}
\end{center}
\vspace{-7mm}
\caption{\textcolor{black}{Overview of IDNet: We have 5979 face portrait photos, used to create 5979 distinct samples for each document type. For each such sample, we further create six fraud samples with different fraud patterns.}}
\label{fig:overview}
\end{figure}

To address these limitations and challenges, this paper has made the following unique contributions:

\noindent
$\bullet$ \textbf{We created and open-sourced a novel {synthetically} generated identity document dataset called IDNet (Sec.~\ref{sec:dataset})}, which contains \textcolor{black}{$837,060$} documents from $20$ types, as illustrated in Fig.~\ref{fig:overview}. Each document type has $5,979$\footnote{{We have $5,979$ distinct synthetic portrait photos that are available for this study, as explained in Sec.~\ref{sec:methodology}.}} distinct document samples that can serve as ground truth for forging detection (i.e., forging modifies part of the document, while counterfeiting attempts to generate a fake document~\cite{aamva1}). \textcolor{black}{For each document sample, we generate \textcolor{black}{six} forged samples with different fraud patterns. The first two fraud patterns are included in existing datasets~\cite{boned2024synthetic}, which are (1) Crop and Move: Cropping a random field from one identity document and moving it to another document; and (2) Impaint and Rewrite: Inpainting a random field and replacing the text in the field by using a different font style or size. We then created four patterns that haven't been implemented in any publicly available datasets: (1) face morphing, (2) portrait substitution by a disqualified portrait, (3) direct alterations in text fields, including random changes to text content, font, and background color schemes without inpainting, and (4) various combination of these fraud patterns. The selection of these specific fraud patterns is informed by their prevalence in real-world fraud instances~\cite{onfido} and their intersection with personal identifier information (PII) fields, thereby posing a substantial challenge for research in privacy-preserving fraud detection. In total, we have $41,853$ document samples for each document type.} To our knowledge, IDNet is the most comprehensive public identity document dataset involving representative fraud types. 

\noindent
$\bullet$ \textbf{We designed and implemented a novel AI-assisted and cost-effective pipeline and methodology to generate identity documents that satisfy research requirements (Sec.~\ref{sec:methodology}).} The pipeline uses {Stable Diffusion 2.0~\cite{yang2023diffusion}, which is free and open-sourced}, to remove portrait photos and other PII information from publicly available sample identity documents (e.g., released by the Department of Motor Vehicles (DMV)) to create templates for different types of identity documents. All portrait pictures used to fill in the templates are artificially generated using AI ~\cite{generated.photos}. The pipeline includes a large language model  (LLM)~\cite{wei2022emergent}, ChatGPT-3.5-turbo, to generate metadata information to fill into ($9$ to $21$) text fields in each template, such as name, address, and DOB, based on the age, {sex}, and ethnicity group associated with the photo. \textcolor{black}{The filling process will automatically search for font size, style, color, and filling coordinates for each field guided by Bayesian optimization.}  {ChatGPT-3.5-turbo achieved similar performance with ChatGPT-4 on our task but required a significantly lower cost ($\$1.5$ vs. $\$60$ for generating $1$ million tokens).} The generative models selected in the pipeline meet the task requirements while satisfying the privacy, latency, and cost constraints. 
%
The entire pipeline can produce an identity document in $0.14$ second on a server with dual Intel Xeon Gold 6226 24-core CPU processors, four Nvidia GeForce 2080 Ti GPUs, and $196$ GB memory. Each document can be produced at an operational cost less than $\$0.0001$. Our approach prioritizes the generation of identity documents that, while not intended for illicit use, are sufficiently authentic to support research demands. We evaluated the metadata diversity, the document fidelity (similarity to real-world documents), the fraud stealthiness (similarity to documents without frauds), and task utility for IDNet in Sec.~\ref{sec:quality}. 
%



\noindent
$\bullet$ \textbf{We conducted evaluations to illustrate the use scenarios of the IDNet benchmark (Sec.~\ref{sec:cases})}. We first explored a practical application of our dataset within a privacy-preserving framework, {showcasing} how new opportunities and challenges arise with our new dataset. We focus on two standard privacy-preserving techniques: \textit{masking}, which involves obscuring sensitive information; and \textit{Pixel-DP}~\cite{lecuyer2019certified}, where pixel-level perturbation based on differential privacy is applied to entire images. Our evaluation centers on assessing the impact of these techniques on fraud detection performance within our privacy-protected data. We observed a notable reduction in the effectiveness of fraud detection when employing the existing privacy-preserving methods we tested. {This key finding illuminates the inherent challenges in designing privacy-preserving algorithms to balance accuracy and privacy.} We also compared the accuracy of various face morphing detection algorithms, which showed that IDNet can serve as a benchmark for specific-purpose fraud detection algorithms, such as face morphing detection. We further used IDNet to investigate cross-type fraud detection (i.e., between two different document types, e.g., Arizona Driver License vs. Spain Passport) and identified an interesting research gap that fraud detection models trained on one type of document {may not} generalize to other types of documents, which argues for new techniques to train cross-type fraud detection models.  
Furthermore, we demonstrated IDNet's diverse document types enable it to serve as a benchmark for evaluating data integration algorithms. For example, we showcased an LLM-based identity document transformation pipeline that automatically converts various IDNet identity documents into a standardized schema.
\textcolor{black}{Most importantly, we showed that IDNet can be used as a foundation to efficiently create a large-scale synthetic identity document dataset within various camera/video capturing environments, e.g., captured by different mobile devices, with different indoor/outdoor backgrounds, and under different lighting conditions.}
%

Before delving into the specifics of our technical contributions, this paper starts with a detailed survey (Sec.~\ref{sec:survey}) about (1) the security features and {personally identifying} information (PII) that exist in various identity documents and (2) the existing publicly available identity document datasets.

\section{Background}
\label{sec:survey}
\subsection{Identity Documents}

As illustrated in Fig.~\ref{fig:overview-AZ-DL}, an identity document usually contains (1) security features, (2) PII information, and (3) other information, which are explained as follows:

\begin{figure}[!htbp]
\begin{center}
  \includegraphics[width=0.9\linewidth]{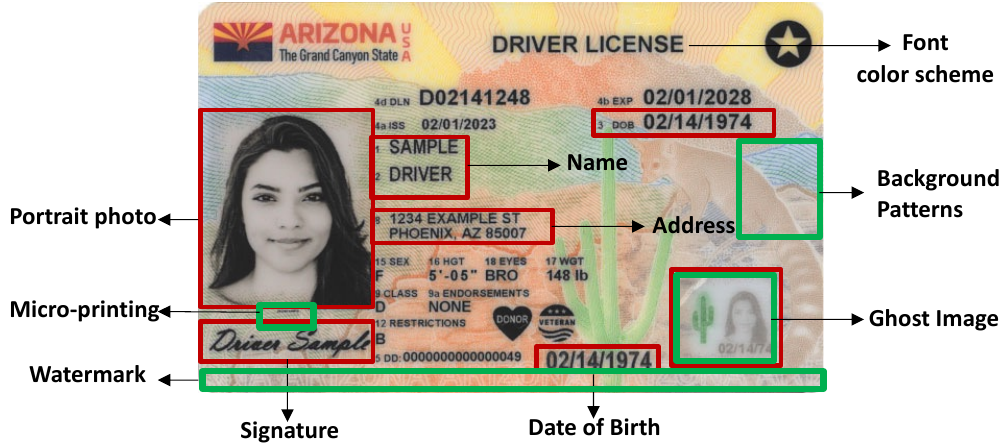}
\end{center}
\vspace{-2mm}
\caption{Overview of Identity Documents -- Taking Arizona Drivers' License Card as an example. Personal identifier information is highlighted in {\textcolor{red}{red}} rectangles, and some example security features are highlighted in {\textcolor{green}{green}} rectangles.
}
\label{fig:overview-AZ-DL}
\end{figure}

\noindent
\textbf{Security Features} {that we explore in this work focus on those that are amenable to digital capture and analysis under standard white lighting conditions, including barcodes, watermarks, micro-printing, guilloche (also known as rainbow printing), distinct color schemes, unique text font, barcodes, and the machine-readable zones (MRZ).}
%

\noindent
\textbf{{Personally Identifiable} Information (PII)} {includes but are not limited to} the portrait photo, signature, barcode, family name, given name, DOB, customer identifier, cardholder address, and ghost image. First, when generating the IDNet datasets, we must not disclose any PII information from the real world. Second, given the {need} to develop new privacy-preserving methods that prevent fraud detection or other analysis processes from disclosing PII information, a primary goal of designing our novel identity document benchmark dataset is to facilitate such analysis by providing fraud patterns that overlap with PII fields and pose challenges for privacy-preserving fraud analysis.

\noindent
\textbf{Other Information}
includes but is not limited to date of issue, date of expiry, document discriminator, endorsement, restrictions, date of first issue, separate expiry, name suffix, weight, height, {sex}, endorsement, restrictions, 
etc. These fields do not contain any personal information and can be disclosed during the analysis process.

\subsection{A Survey of Existing Public Identity Document Datasets}


{Existing} publicly available document datasets are designed for recognizing, classifying, and restoring information from documents captured as videos or photos using mobile devices. 
For example, the SmartDoc dataset~\cite{chazalon2017smartdoc} is a publicly available document dataset. It contains a training set of $10$ document samples captured as video clips (i.e., document capture samples) by a video camera. Each training sample also contains an image of the document used to produce the sample, which is considered the ground truth for comparison to the document restored from the video. In addition, SmartDoc offers a testing set of $37$ document capture samples. This dataset includes only a few identity documents.
The LRDE identity document image database~\cite{ngoc2018saliency} comprises $100$ videos for a dozen different types of visas and passports from various countries using different environmental conditions and several kinds of smartphones.

MIDV-500~\cite{arlazarov2019midv} is a publicly available identity document dataset that contains $500$ video clips. These video clips are produced from $50$ different identity document types, including $17$ types of ID cards, $14$ types of passports, $13$ types of driving licenses, and $6$ other identity documents of various countries. Each of the $50$ document types will be used in $5$ different backgrounds to generate $10$ video clips that last at least $3$ seconds in duration using two mobile phone devices. MIDV-500 aims at facilitating simple analysis tasks like face detection, optical character recognition (OCR), and document type classification. MIDV-2019~\cite{bulatov2020midv} extended the MIDV-500 dataset to include four more videos with distorted identity documents and different lighting conditions for each identity document type.  MIDV-2020~\cite{bulatovich2022midv} increased the number of unique document samples to $1000$ ($100$ unique documents for each of $10$ document types). 

Datasets featuring fraudulent identity documents remain scarce.
FMIDV~\cite{al2023guilloche} addresses this gap by introducing seven forged IDs for each sample in the MIDV-2020 dataset, focusing on guilloche-pattern fraud. To generate a forged ID, they randomly selected a few blocks that only contained guilloche patterns from one ID and copied these blocks to random locations in the blank area of another ID. FMIDV is limited to the single guilloche-related copy-and-move fraud pattern and overlooks many popular identity document fraud patterns. SIDTD~\cite{boned2024synthetic} is the most recent extension of the MIDV-2020 dataset. It used crop-and-move and inpainting techniques to create simple frauds, containing $1222$ fraud documents. Crop-and-Move will copy a text field from one document to another document. Inpainting will change the font style of the texts in a selected field, but will not change the text contents. Yet, the MIDV family has a limited number of distinct document samples, ranging from $50$ to $1000$, which poses challenges for AI/ML applications in achieving high accuracy.

The Brazilian Identity Document Dataset (BID)~\cite{de2020bid} also focuses on simple tasks such as automatic text extraction, OCR, and document classification rather than fraud detection. The BID dataset contains $28,800$ document images from eight different document types.  These documents are created by altering authentic documents, which introduces discrepancies between the dataset and genuine documents, notably in the blurring of portrait photos, limiting its applicability in fraud detection involving portraits.

\section{The IDNet Benchmark Dataset}
\label{sec:dataset}

Leveraging a cutting-edge AI-assisted pipeline, outlined in Sec.~\ref{sec:methodology}, we have developed an identity document dataset, named \textit{IDNet}, which is entirely synthetically generated and devoid of any private information. This dataset encompasses a total of \textcolor{black}{$837,060$} identity documents, spanning across $20$ different document types. For each document type, there are $5,979$ unique document samples. Each sample comprises one authentic copy alongside six fraudulent variations, which include face morphing, portrait substitution, text alteration, a combination of these fraudulent techniques, and two fraud patterns (inpaint-and-rewrite, and crop-and-move) leveraged from the SIDTD benchmark,  as detailed in Sec.~\ref{sec:fraud}. IDNet stands as the largest publicly accessible identity document dataset to date, as depicted in Fig.~\ref{fig:overview}. Detailed statistics and Zenodo URLs for each document type within IDNet are systematically presented in Tab.~\ref{tab:type-statistics}.

\begin{table}[]
    \centering
    \vspace{-10pt}
    \caption{Overview of $20$ types of identity documents in IDNet }
    \begin{tabularx}{\textwidth}{c *{5}{>{\centering\arraybackslash}X}}
    \toprule
     & \multicolumn{5}{c}{Statistics}\\
    \cmidrule(lr){2-6}
    country/area & type & \#text fields & \#image fields & storage size &url \\
    \midrule
     Arizona (US) & DL & 17 & 3 & 17GB & \href{https://zenodo.org/records/10573853}{URL}\\
     California (US) & DL & 18 & 3 & 25GB & \href{https://zenodo.org/records/10570622}{URL}\\
     Nevada (US) & ID card & 15 & 3 & 14GB & \href{https://zenodo.org/records/10574073}{URL}\\
     North Carolina (US) & DL & 17 & 5 & 50GB & \href{https://zenodo.org/records/10573853}{URL}\\
     Pennsylvania (US) & DL & 16 & 3 & 56GB & \href{https://zenodo.org/records/10574012}{URL}\\
     South Dakota (US) & DL & 16 & 3 & 32GB  &\href{https://zenodo.org/records/10574172}{URL}\\
     Utah (US) & DL & 21 & 3 & 35GB & \href{https://zenodo.org/records/10574172}{URL}\\
     Washington D.C. (US) & DL& 16 & 3 & 21GB& \href{https://zenodo.org/records/10574215}{URL}\\
     West Virginia (US) & DL & 17 & 3 & 41GB & \href{https://zenodo.org/records/10574215}{URL}\\
     Wisconsin (US) & DL & 13 & 4 &50GB& \href{https://zenodo.org/records/10574073}{URL}\\
     Albania & Passport & 11 &3 & 27GB & \href{https://zenodo.org/records/10611634}{URL}\\
     Azerbaijan & Passport & 18 & 3 & 17GB & \href{https://zenodo.org/uploads/10602369}{URL}\\
     Estonia & ID card & 10 & 3& 20GB&\href{https://zenodo.org/records/10611634}{URL}\\
     Finland & ID card & 9 & 4 & 6GB & \href{https://zenodo.org/uploads/10602369}{URL}\\
     Greece & Passport & 20 & 1 & 8GB & \href{https://zenodo.org/uploads/10602369}{URL}\\
     Latvia & Passport & 20 &4 & 24GB & \href{https://zenodo.org/records/10570622}{URL}\\
     Russia & Passport & 9 & 1&  8GB& \href{https://zenodo.org/records/10570622}{URL}\\
     Serbia & Passport & 20 & 3 & 34GB & \href{https://zenodo.org/uploads/10602369}{URL}\\
     Slovakia & ID card & 10 & 2 & 6GB &\href{https://zenodo.org/records/10570622}{URL}\\
     Spain & ID card & 10 & 2 & 7GB &\href{https://zenodo.org/records/10611634}{URL}\\
    \bottomrule
    \end{tabularx}
       \vspace{-10pt}
    \label{tab:type-statistics}
\end{table}

The IDNet {dataset} is released as identity document image files along with JSON files that describe the metadata of each document. The metadata for each positive sample (w/o fraud patterns) includes the document identifier, face image ID, name, {sex}, date-of-birth (DOB), issuing date, and expiration date.  The metadata for each fraud identity document contains additional information, such as the fraud type and fraud parameters. For portrait substitution, we recorded the identifier of the original face, and the identifier of the new face. For face morphing~\cite{venkatesh2021face}, we documented the ID and the morphing weight of each morphed face~\cite{ngan2022face}. 
For text field replacement, we stored information such as the original and updated text content, font style, size, and color scheme.

The IDNet dataset is accessible publicly across various Zenodo repositories, with direct links listed in Tab.~\ref{tab:type-statistics}. \eat{We also open-sourced the code for generating the dataset \url{https://github.com/asu-cactus/fake_id_synthesis.git}.}

It is crucial to understand that the primary aim of this research is not to replicate real-world identity documents exactly. Instead, our goal is to furnish a rich and varied collection of documents that adhere to identity document design norms. These documents are intended to support the research community in conducting privacy-conscious document analysis and fraud detection. The specific aims include:

\noindent
$\bullet$ \textbf{Diversity.} We aim for the dataset's document metadata diversity (e.g., entropy) to meet or exceed that of other synthetic benchmarks like MIDV-2020.  The evaluation of this attribute will be discussed in detail in Sec.~\ref{sec:diversity}.

\noindent
$\bullet$ \textbf{Fidelity.} The visual similarity of IDNet's document images to their real-world counterparts should be high, aiming for a Structural Similarity Index (SSIM) greater than 0.85. This aspect will be assessed in Sec.~\ref{sec:fidelity}.

\noindent
$\bullet$ \textbf{Stealthiness.} 
To maintain the stealthiness of the fraudulent modifications, the structural similarity between documents with and without fraudulent modifications should be nearly indistinguishable, with an SSIM greater than 0.95. This criterion will be explored in Sec.~\ref{sec:stealthiness}


\noindent
$\bullet$ \textbf{Utility.} The dataset should enable AI/ML models to achieve accuracy levels comparable to those attained with real-world datasets. The utility of IDNet in the context of AI/ML tasks will be examined in Sec.~\ref{sec:morphing-utility} \textcolor{black}{and Sec.~\ref{sec:text-replacement-utility}. More evaluations of various use cases can be found in Sec.~\ref{sec:cases}}.

Before presenting the evaluation of IDNet's quality, it is crucial to understand the methodology employed in the dataset's generation.

\section{Generation of the Identity Document Dataset }
\label{sec:methodology}

\subsection{Template Generation based on Image Diffusion Model}
\begin{figure}[!htbp]
\vspace{-5mm}
\begin{center}
  \includegraphics[width=\linewidth]{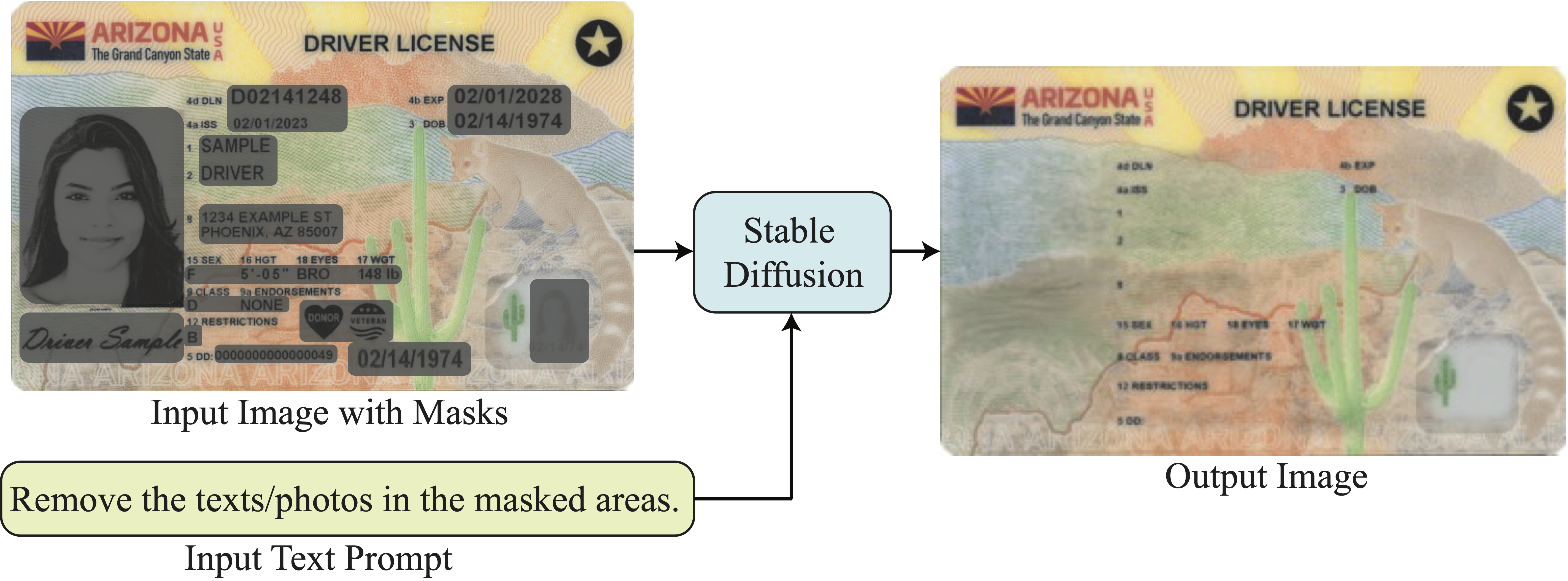}
\end{center}
\vspace{-5mm}
\caption{Illustration of the identity document template generation process.}
\label{fig:template_generation}
\end{figure}

To create {the} IDNet dataset, it is essential first to acquire a template for each type of identity document. However, high-quality, blank templates of real-world documents are generally unavailable. To overcome this, we utilize image generative models, such as diffusion models~\cite{ho2020denoising}\cite{rombach2022high}\cite{song2020denoising}, to produce our ID templates by erasing the content from actual identity documents. Specifically, we employed the Stable Diffusion version 2.0 from Hugging Face~\footnote{\url{https://huggingface.co/stabilityai/stable-diffusion-2}}, which is based on the Latent Diffusion Model~\cite{rombach2022high}, to create templates from $20$ different types of real-world IDs. This model is adept at editing masked areas of an input image in accordance with text prompts. For our purposes, we mask all customizable information on the IDs as illustrated in Fig.~\ref{fig:template_generation} and direct Stable Diffusion with the prompt "remove all texts/photos in the masked areas." Consequently, the model adeptly eliminates customized data from the document and replenishes the masked sections with appropriate backgrounds. We demonstrate this template generation technique in Fig.~\ref{fig:template_generation}, showcasing the Arizona Driver's License as an illustrative example.

In instances involving complex identity documents, such as the passport from the Republic of Azerbaijan, where the information to be removed intersects with elements we wish to retain, the model faces challenges in achieving desirable outcomes without precise mask adjustments. To enhance the quality of the generated templates in these scenarios, we engage in an iterative process of applying the diffusion model with progressively refined masks and directives until a satisfactory template is achieved.

\begin{figure}[!htbp]
\vspace{-10mm}
\begin{center}
  \includegraphics[width=\linewidth]{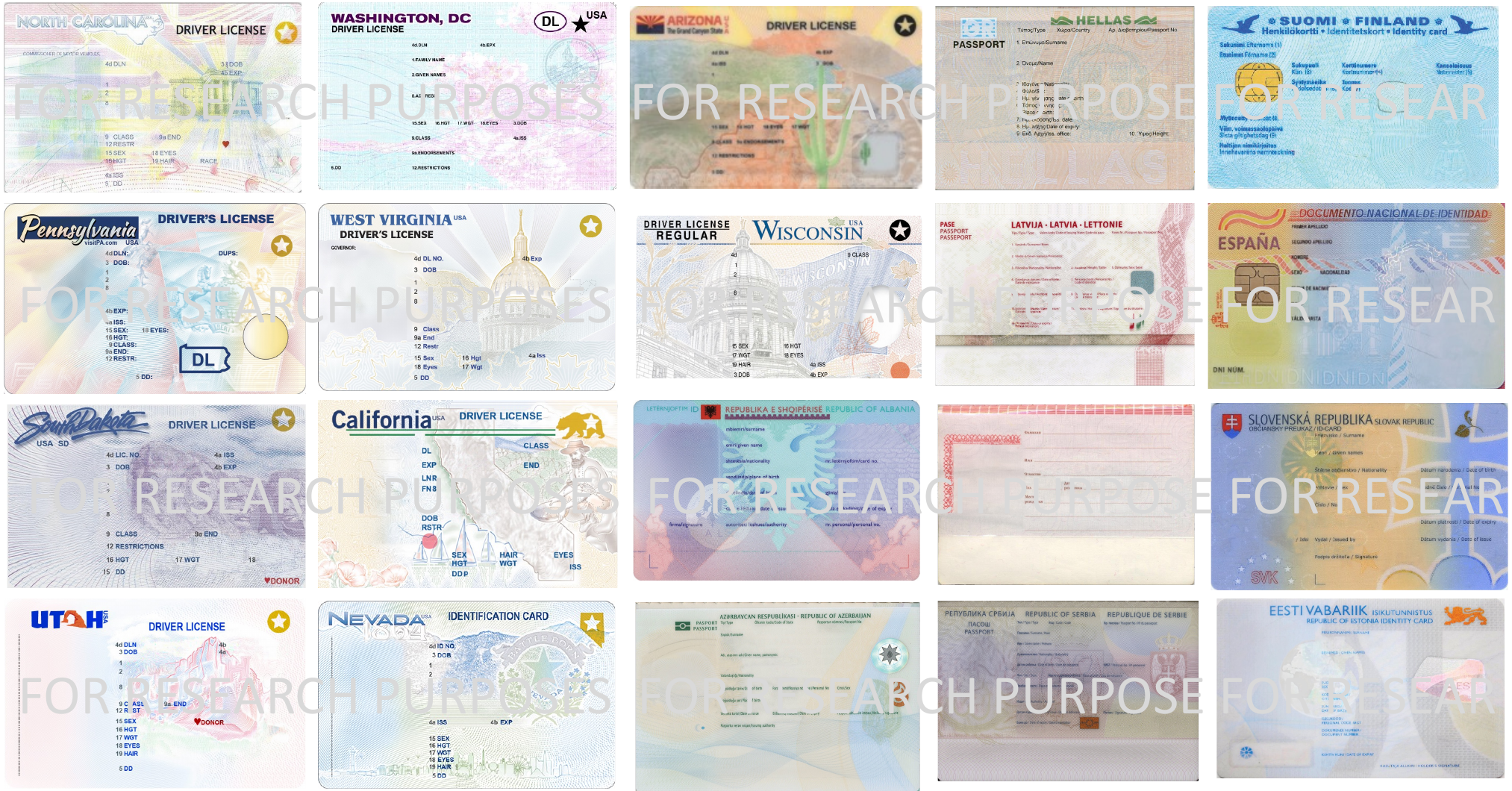}
\end{center}
\vspace{-5mm}
\caption{Illustration of the generated identity document templates (We added texts such as "For Research Purposes" to avoid abuse uses).}
\label{fig:templates}
\end{figure}

\subsection{Metadata Information Generation}
\label{sec:generation}

To synthesize the information to be filled into the identity document templates, several key factors were considered. First, 
some fields correlate with each other. For example, the sex, ethnicity group~\cite{ethnicity-groups}, age, and eye color should match with the portrait photo and the name should match the sex and ethnicity group. Second, different identity cards may have different requirements over ages, portrait photos' facing directions and facial expressions, and so on. For example, many US states have a minimum age limit for applying for driver's licenses. To address these issues, we used a well-known public face image dataset for academic research\footnote{https://generated.photos/}, which is also used by all benchmarks in the MIDV family. All faces in the dataset are {synthetically} generated. The total dataset consists of $10,000$ synthetic human face images with metadata such as face landmarks, sex, emotion, ethnicity, eye color, age, facial expression, etc. {Not all $10,000$ photos are qualified for ID documents.} To meet the age and face image requirements of many identity documents, the photos are filtered based on the following characteristics: {age, wearings, facial expression, and headpose. To alleviate the manual burden to filter photos, the metadata are used to filter qualified photos from disqualified photos:}(1) Age less than $18$; (2) Probability of emotion "neutral" or "happiness" less than $0.8$; (3) Any of the head pose attributes, i.e., "pitch", "roll", or "yaw", are greater than $9$. 
{After the first round filtering, the two set of photos are manually processed to obtain a final collection of qualified photos.} As a result, $5,979$ photos were extracted from the {full} collection {of synthetic dataset and $4021$ photos are disqualified. Tab.~\ref{tab:filtering} shows the number of disqualified photos of different types. Miscellaneous type include photos with obvious artifacts or with persons wearing sunglasses or head covering.} We also preprocessed the portrait photos to meet the specific requirements for photo sizes and background colors for different types of documents. 

\begin{table}
\small
\centering
\caption{{Number of disqualified photos of different types}}
\label{tab:filtering}
\begin{tabular}{ |p{2.8cm}|p{2.0cm}|p{2.5cm}|p{2.0cm}|p{2.0cm}| } 
 \hline
  Disqualification types & Age & Facial expression & Head pose & Miscellaneous \\ \hline
Number of photos& 661 & 77 & 3232 & 51\\ \hline
\end{tabular}
\end{table}

Then, taking the document {type} of Arizona State Driver's License as an example, we generate metadata information using random generators and Large Language Model (LLM) such as OpenAI's ChatGPT API (i.e., we used ChatGPT version 3.5 turbo API).  

\noindent
$\bullet$ \textbf{The sex information} was provided in the metadata of any face image, which is in the form of two numbers representing the probability of being a male and a female. We chose the sex by sampling the probability distribution. 

\noindent
$\bullet$ \textbf{The eye color information} is generated similarly. We used the probability distribution of eye colors specified in the metadata of a face image and chose the eye color by sampling the probability distribution.

\noindent
$\bullet$ \textbf{The height information} was generated randomly based on heuristic distributions. We first uniformly sampled a number in the range of $[0,1)$. If this number was less than $0.6$, we chose the height to be $5$ feet. For a female, if this number was greater than $0.6$ but less than $0.9$, the height was set to $6$ feet; if this number was greater than $0.9$, the height was set to $7$ feet. For a male, if this number was greater than $0.6$ but less than $0.8$, the height was set to $6$ feet; if this number is greater than $0.8$, the height was set to $7$ feet. For both males and females, an integer was uniformly sampled from the range $[0,12)$ to form the inch part of height.

\noindent
$\bullet$ \textbf{The weight generation} was sampled from the range of $[100,200]$ for female, and the range of $[100,250]$ for male, based on the generated height information.

\noindent
$\bullet$ \textbf{The document discriminator (DD)}, which uniquely identifies a particular driver's license or ID card,  is composed of digits and letters with a total length of $16$~\cite{aamva1}. We used an algorithm to randomly sample the number of letters, the position of the letters, and the remaining digits.

\noindent
$\bullet$ \textbf {The driver's license number (DLN)} was generated by sampling a number from the range of $[0,100,000,000)$ without replacement for each of $5,979$ identity documents. For certain types of documents, where DLN was alphanumeric, e.g. Virginia driver’s license, we adopted an approach similar to generating the document discriminator.

\noindent
$\bullet$ \textbf{The date of birth} was generated based on the age information associated with the portrait photo. We extracted age from the photo's metadata and subtract the current year by age to obtain the birth year. The month is uniformly sampled from twelve months, and then the date was uniformly sampled from the days of the month. 

\noindent
$\bullet$ \textbf{The issuing date and expiration date} were generated based on the document validity period. Supposing the validity period is five years, we first uniformly sampled an integer from $0$ to $5$, then we subtract the current year by this integer as the year of the issue date. The year of the expiration date was set to the year of the issue date plus $5$. The month and date were sampled randomly.

\noindent
$\bullet$ \textbf{The driver license class} was randomly sampled following a heuristic distribution. We assumed class D is the most common class. We first uniformly sampled a number in the range of $[0,1)$. If the number was greater than $0.9$, we uniformly sampled a class from A, B, and C. Otherwise, we chose class D.

\noindent
$\bullet$ \textbf{The first and last names} were generated using an LLM. First names were generated by querying the ChatGPT-3.5 API with the following prompt, "Please generate 50 distinct English first names for {ethnicity} {sex}". Here, the ethnicity and sex in curly braces were replaced by the ethnicity and sex variables extracted from the metadata of the corresponding portrait photo. Last names are generated with the following prompt, "Please generate 50 distinct English last names for {ethnicity} families". The full names were randomly selected based on the ethnicity group associated with the portrait photo. For each identity document, the names were drawn from all full name combinations without replacement. For European countries, names were generated similarly in the corresponding language. For example, to generate first names for Spanish ID documents, the following prompt was sent to ChatGPT-3.5, "Please list $40$ Spanish male given names and $40$ female given names in uppercase." Last names were generated with the following prompt "Please list $50$ Spanish surnames in uppercase."

\noindent
$\bullet$ \textbf{Addresses} were also generated using ChatGPT-3.5. To guide the model to generate high-quality addresses, the following prompt was used: "Please generate $50$ fictional Arizona addresses. First, the city name was generated based on the state and postal code. Second, a street name was generated in the previously generated city. Third, a random number was generated with one,  two, three, or four digits as the street number. Lastly, combine the above information in the format "{street number} {street name}, {city}, {state} {postal code}". We found that ChatGPT-3.5 struggled to generate diverse addresses. Therefore, these $50$ generated addresses were used as seeds and augmented by replacing the street numbers with a random number produced by a Python script.

{\color{black}{
\subsection{Add Generated Information to the Identity Document Templates}
Once all the synthetic information was generated, it was added to the corresponding template. One of the challenges involved the selection of font size and style that closely approximates those found on genuine documents. 

{To optimize the generation of text overlays on genuine documents, we applied Bayesian optimization~\cite{frazier2018tutorial} to identify the best parameters that make the generated document as similar to the original as possible. The aim was to maximize the Structural Similarity Index (SSIM)~\cite{hore2010image}, a metric that quantifies the visual similarity between two images by considering luminance, contrast, and structure. SSIM provides a robust measure for evaluating the quality of the generated image, ensuring it closely matches the original in terms of appearance and detail.}

{In our optimization process, we divided each sample into several segments, as illustrated in Fig.~\ref{fig:segment-based-search}. For each segment, we focused on tuning several critical parameters (font style, font size, font color, font width, and position) to achieve the desired outcome. Bayesian optimization was employed to explore the parameter space efficiently. The process iteratively evaluated combinations of parameters, adjusting them to maximize SSIM. This approach allowed us to systematically and effectively enhance the visual similarity between the generated and genuine documents.}

\begin{figure}[!htbp]
\begin{center}
  \includegraphics[width=0.5\linewidth]{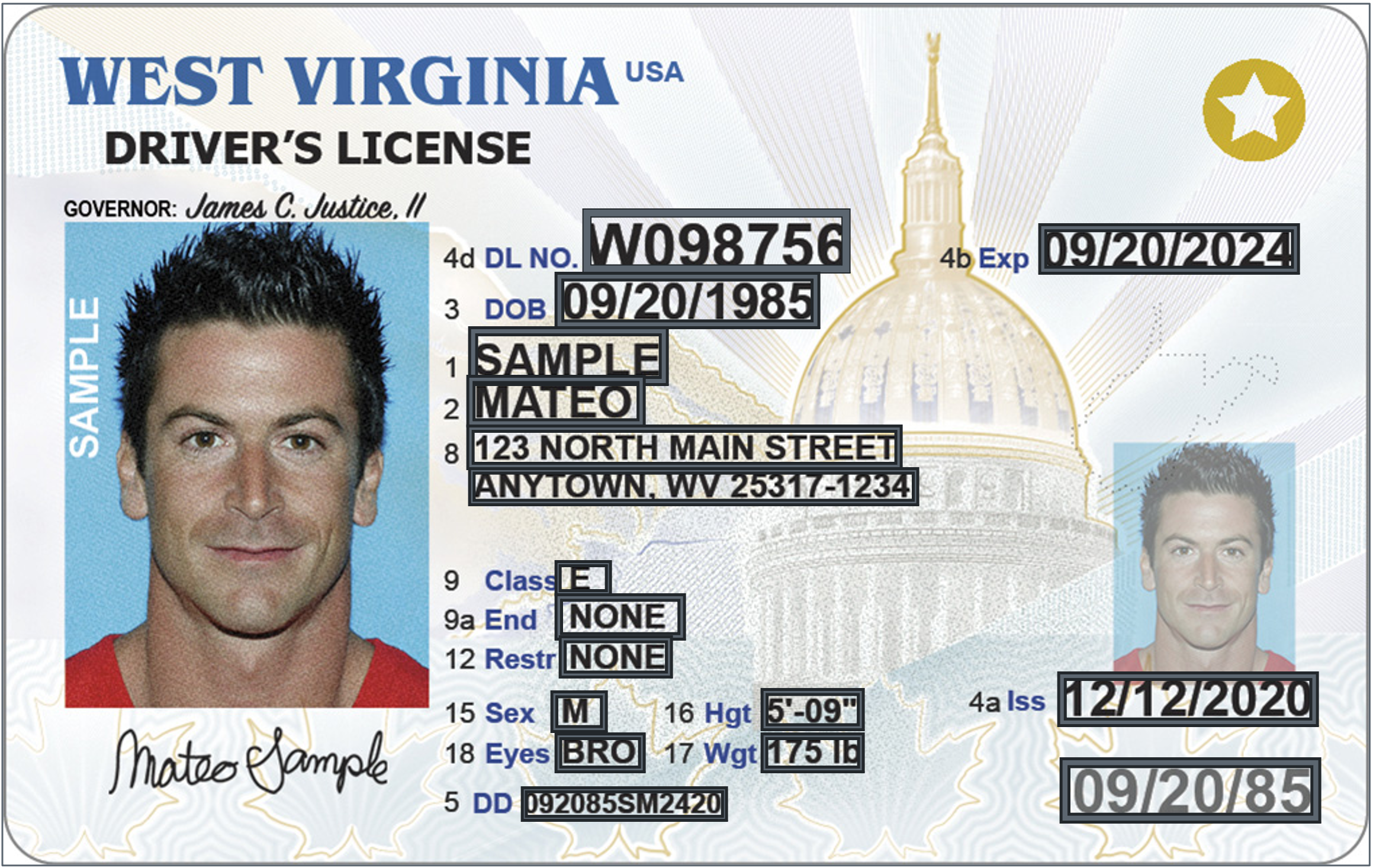}
\end{center}
\vspace{-2mm}
\caption{{Illustration of segment-based parameter search}
}
\label{fig:segment-based-search}
\end{figure}

{We also compared the results without Bayesian optimization (BO), for which we manually tuned parameters, and the results are shown in Tab.~\ref{tab:Bayesian-result}. We can see that BO brought a significant improvement in SSIM.}

\begin{table}
\small
\centering
\caption{{Comparison of results with and without Bayesian optimization(BO) method}}
\label{tab:Bayesian-result}
\begin{tabular}{ |p{2.2cm}|p{2.2cm}|p{2.2cm}|p{2.2cm}| } 
 \hline
  ID Type & SSIM w/o BO & SSIM w/ BO & Improved(\%) \\ \hline
    WV& 0.881062 & 0.967402 & 9.80\% \\ \hline
    AZ& 0.883343 & 0.956468 & 8.28\% \\ \hline
    DC& 0.922293 & 0.958049 & 3.88\% \\ \hline
    CA& 0.920069 & 0.975129 & 5.98\% \\ \hline
\end{tabular}
\end{table}

{By using Bayesian optimization, we leveraged a probabilistic framework that adapts to the complexities of the parameter space, providing a scalable and efficient method for finding the optimal settings for text overlay. We also measured the converging rate of the Bayesian optimization. As shown in Tab.~\ref{tab:converging-rate}, We can see the convergence speed is very fast and when the step is 100, we can get very good result.}

\begin{table}
\small
\centering
\caption{{Convergiing Rate of Bayesian Optimization}}
\label{tab:converging-rate}
\begin{tabular}{ |p{2.8cm}|p{3.0cm}|p{2.5cm}| } 
 \hline
  Number of steps & SSIM(One Segment) & SSIM(Full Image) \\ \hline
    0& 0.124484 & 0.881062 \\ \hline
    30& 0.687587 & 0.948555 \\ \hline
    100& 0.895166 & 0.955958 \\ \hline
    200& 0.898774 & 0.957535 \\ \hline
    500& 0.901406 & 0.958891 \\ \hline
\end{tabular}
\end{table}


In addition to the textual PII information and the portrait photo, many types of identity documents (IDs) encompass the ghost image (serving both as a security feature and additional PII) and the signature of the ID holder. 

The generation process for the ghost image commences with the removal of the portrait image's background, followed by its conversion into a single-channel image retaining only the luminosity level data. This transformation effectively shifts the image to grayscale, closely mirroring the visual characteristics of a genuine ID's ghost image. The resultant ghost image is then seamlessly integrated into the template.

A specialized randomizer algorithm was developed for signature generation. It first hashes the ID holder name into a reasonably compact string, which is subsequently applied to the template using a font randomly selected from fourteen fonts open-sourced under permissive licenses, which emulate various handwriting styles. This dual process of name hashing and font randomization ensures a varied and authentic representation of individual handwriting styles across the dataset.
}}

\subsection{Fraud Patterns}
\label{sec:fraud}
\begin{figure}[!htbp]
\begin{center}
  \includegraphics[width=\linewidth]{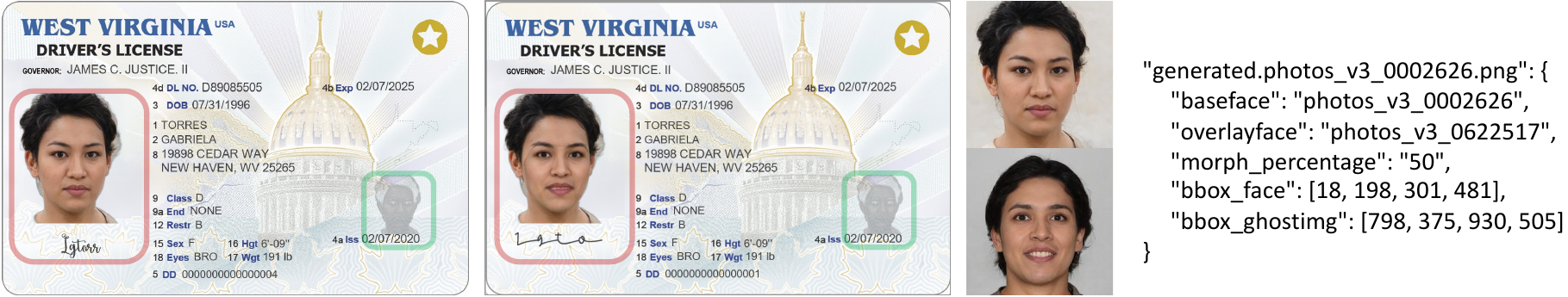}
\end{center}
\vspace{-2mm}
\caption{Illustration of Portrait Morphing Fraud
}
\label{fig:portrait-morphing-fraud}
\end{figure}

Although our generated dataset is a {synthetic} dataset, we can consider it a "representative" dataset in research environments given its utility in many research tasks as detailed in Sec.~\ref{sec:quality} and Sec.~\ref{sec:cases}. Therefore, it is reasonable to create forged identity documents on top of it. Subsequently, we enumerated and analyzed various prevalent fraud patterns observed in contemporary forged IDs. A pivotal component of our investigative methodology involved the generation of a batch of forged IDs, incorporating one or more of these identified patterns. The fraud patterns delineated in this study are described below:

\vspace{3pt}
\noindent
$\bullet$ \textbf{Face-Morphing Fraud}~\cite{venkatesh2021face}~\cite{korshunov2013using}:
recently emerged as a notable threat~\cite{onfido}. This type of fraud leverages the natural variations in human facial features over time, creating opportunities for identity deception. It operates on the premise that an individual's facial characteristics can significantly alter from those documented on their official identification. This variance enables an attacker (referred to as Person A) to misuse the identification of another individual (Person B), provided there is a sufficient resemblance between their facial features~\cite{venkatesh2021face}~\cite{morphing-biometrics}. To integrate this complex fraud pattern into our analysis, we adopted cutting-edge image fusion methods {including Image Warp and Cross Dissolve}~\cite{wolberg1998image}. Image wrap aligned key facial landmarks (e.g., eyes, nose, mouth, etc) of one face with those of the other, which ensures that the subsequent blending of the images appears natural and seamless. After the facial images were warped and aligned, Cross Dissolve blended them into a single cohesive image. The outcome is a synthesized facial image that encapsulates the likeness of both input face images concurrently, attaining a level of confidence significant enough to pass visual scrutiny as authentic. To achieve high-quality morphing as suggested by the NIST face morphing report~\cite{ngan2022face}, only the face area of the facial images was averaged after alignment and feature warping. In addition, the face area was adjusted to the face color histogram of the first input facial image. For each of the $5979$ artificially generated photos, we morphed it with another randomly selected photo of the same ethnicity and sex, as illustrated in Fig.~\ref{fig:portrait-morphing-fraud}. We set the blending factor as $0.5$ in the face morphing process following the NIST face morphing tie-2 implementation~\cite{ngan2022face}, which suggests that both faces contribute equally to the morphed face.

\begin{figure}[!htbp]
\begin{center}
  \includegraphics[width=\linewidth]{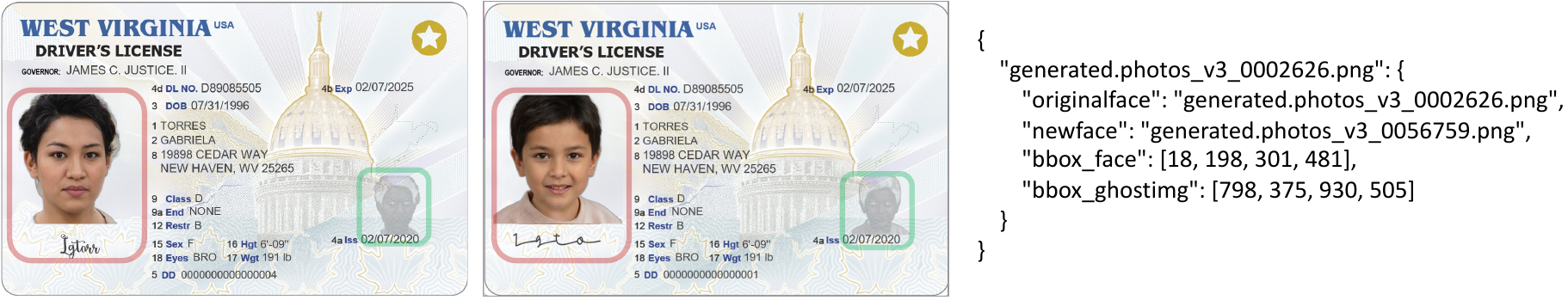}
\end{center}
\vspace{-2mm}
\caption{Illustration of Portrait Replacement Fraud
}
\label{fig:portrait-replacement-fraud}
\end{figure}

\noindent
$\bullet$ \textbf{Portrait Substitution Fraud}~\cite{portrait-substitution1}\cite{onfido}\cite{mercer1998document}.Based on industrial studies~\cite{onfido}, a majority of digital identity document attacks in online platforms in 2023 are at low forging costs.  Portrait substitution is one example, which is to use disqualified digital photos, e.g., photos taken using mobile phones or computer cameras and do not meet photo standards required by the corresponding identity document. To implement this type of fraud, for each identity document, we uniformly sampled one photo from the $4,021$ portrait photos identified as disqualified when we preprocessed the $10,000$ synthetic portrait photos (i.e., the preprocessing procedure is described at the beginning of Sec.~\ref{sec:generation}). We then used that sampled photo to replace the original photo to produce a forged identity document, as illustrated in Fig.~\ref{fig:portrait-replacement-fraud}. 

\begin{figure}[!htbp]
    \subfigure[Easy-level]{
        \includegraphics[width=1\columnwidth]{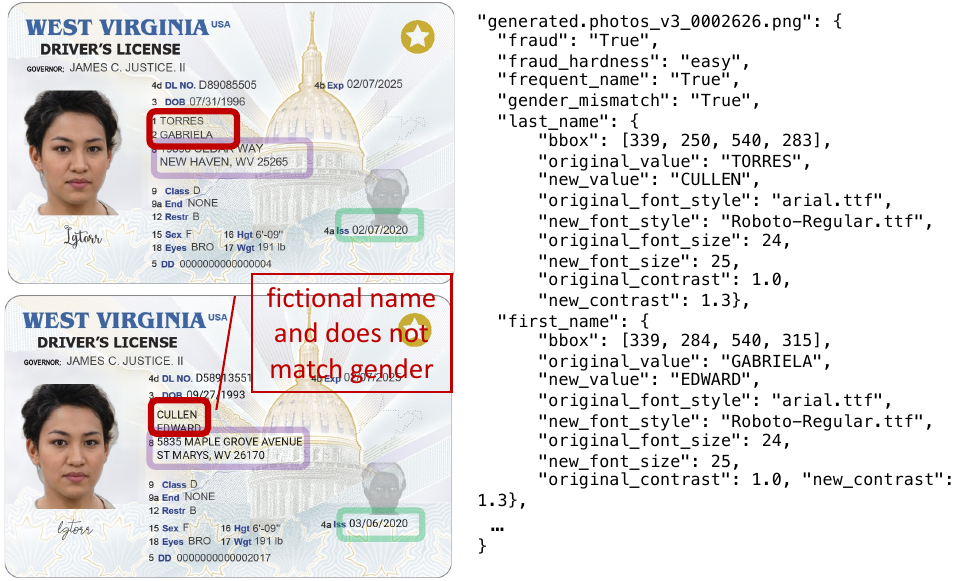}
    }
     \subfigure[Hard-level]{
        \includegraphics[width=1\columnwidth]{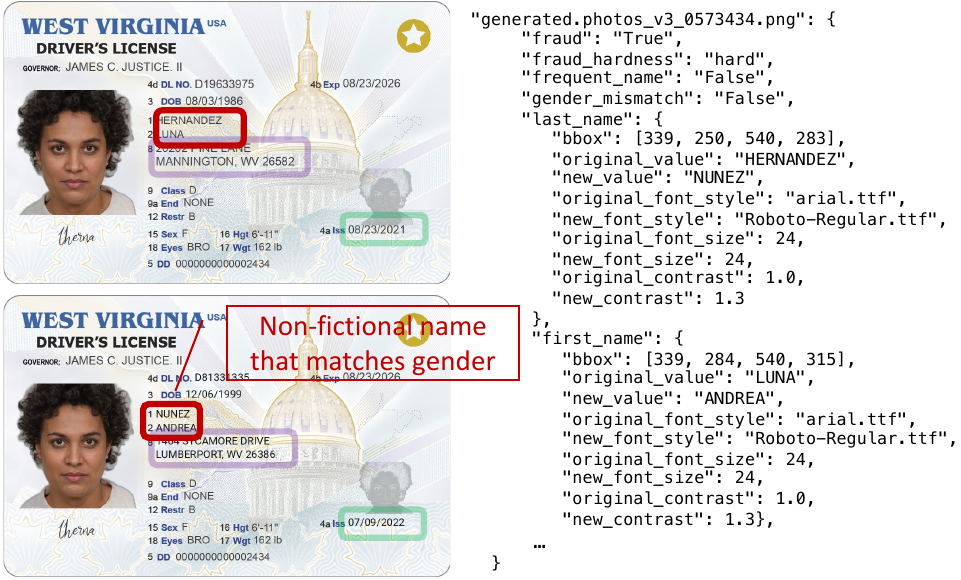}
    }
    \caption{Illustration of Text-Field Replacement Fraud}
     \label{fig:text-replacement-fraud}
\end{figure}

\noindent
$\bullet$ \textbf{Text-Field Replacement Fraud}~\cite{portrait-substitution1}\cite{onfido}. Replacing text field/(s) often leads to subtle alteration in the text font styles, text font size, and background color of PII fields. Counterfeited/forged IDs manipulate specific PII fields, with common targets being the first and last names, sex, DOB, expiration data, etc. To integrate this fraud pattern into our dataset, we randomly changed the font style, font size, and the contrast and saturation of PII fields. This manipulation simulated the changes in the text fields commonly observed in forged IDs~\cite{onfido}. We further classified this fraud into two levels as illustrated in Fig.~\ref{fig:text-replacement-fraud}: (1) easy-level, where the replacing text fields' information (e.g., name, DOB) does not match the portrait photo's sex and age, which is relatively easier to detect; and (2) hard-level, where the forged information is consistent with the rest of the identity document information. The ratio of easy-level fraud was determined to be around $65\%$, and the rest were all hard-level frauds, to be consistent with {a recent industry} survey~\cite{onfido}. Such classification can be leveraged to evaluate anti-fraud methods to ensure they perform comparisons between image and text data, e.g., age verification~\cite{age-verification}.

\vspace{3pt}
\noindent
$\bullet$ \textbf{Mixed Fraud Pattern} For each distinct identity document, we created an additional fraud sample that mixed fraud patterns through the following steps. First, the text replacement frauds was applied to the PII fields. Subsequently, a random variable was sampled to decide whether to apply a face morphing or portrait substitution fraud to the portrait photo of the document. This dataset can be used to test the integration of specialized fraud detection algorithms, each focusing on one fraud pattern.

\begin{figure}[!htbp]
\begin{center}
  \includegraphics[width=\linewidth]{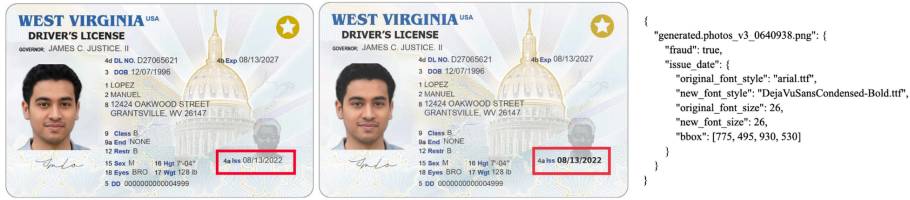}
\end{center}
\vspace{-2mm}
\caption{Illustration of Inpaint and Rewrite Fraud
}
\label{fig:inpaint-and-rewrite}
\end{figure}

\begin{figure}[!htbp]
\begin{center}
  \includegraphics[width=\linewidth]{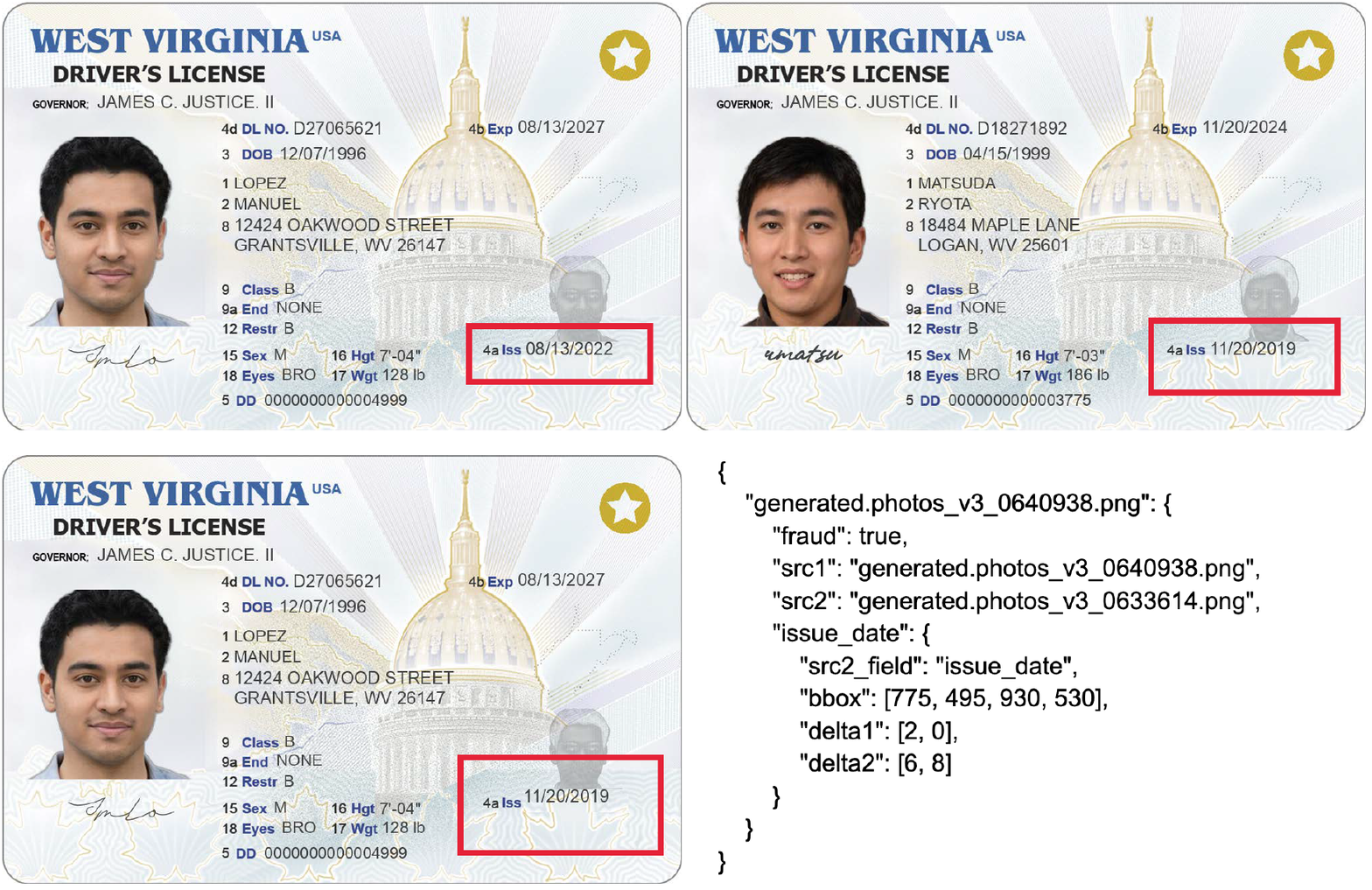}
\end{center}
\vspace{-2mm}
\caption{Illustration of Crop and Replace Fraud
}
\label{fig:crop-and-replace}
\end{figure}

{\color{black}{
\vspace{3pt}
\noindent
$\bullet$ \textbf{Inpaint and Rewrite Fraud Pattern} This fraud technique was utilized in SIDTD \cite{boned2024synthetic} to create counterfeit samples of IDs. It is similar to our earlier described text-field replacement fraud except for several differences: (1) the fraud generation process of this fraud pattern has applied a customized mask to a randomly selected field for inpainting each time; (2) the font style for the replacement text is chosen randomly from the fonts available in SIDTD; and (3) this pattern will not change the background color scheme and text content. 
Figure \ref{fig:inpaint-and-rewrite} illustrates the Inpaint-and-Rewrite pattern as described.

\vspace{3pt}
\noindent
$\bullet$ \textbf{Crop and Replace Fraud Pattern} This fraud pattern, also utilized in SIDTD \cite{boned2024synthetic}, facilitates the exchange of information between IDs of the same class. In this method, PII field is selected randomly from one ID and then cropped and replaced with the PII field of another ID. In both cases, the PII fields are selected randomly, with a $95\%$ probability that the same PII field is chosen in both IDs and a $5\%$ probability that different PII fields are selected.
To ensure a slight variation, a small shift is added when replacing the cropped area of one ID with the other. This shift is randomly selected from a defined range and applied to both the x-axis and y-axis. To avoid perfect alignment between the two IDs, the shift value of 0 is omitted. This ensures there is always a shift, creating a slight distortion in the background due to texture discontinuity. Figure \ref{fig:crop-and-replace} illustrates the Crop-and-Replace pattern as described. 
}}

\subsection{Time and Monetary Costs}

We measured the time and dollars spent at every pipeline stage, as illustrated in Tab.~\ref{tab:costs}. The pipeline ran on a system with dual Intel Xeon Gold 6226 CPUs at 2.70GHz with 24 cores each, four Nvidia GeForce 2080 Ti GPUs, and $196$ GB memory to produce the identity documents. We estimate its cost to be $\$ 2$ per hour based on the costs of AWS EC2 on-demand instances of similar capabilities~\cite{aws-pricing}. The stable diffusion 2.0 model is free and open-sourced. The ChatGPT-3.5-turbo API costs $\$0.5$ for a million input tokens and $\$1.5$ for a million output tokens. We used $1860$ user input tokens and $60,652$ output tokens. While generating the metadata information, we spent $410$ seconds to generate sample names and addresses for $20$ different countries/states and $145$ seconds to derive all metadata information. All generated documents are archived using the free Zenodo service. As a result, the operational cost for producing each identity document is lower than $\$0.0001$ with a latency of $0.14$ second. It demonstrated the cost-effectiveness of the proposed pipeline and the great potential of using the pipeline to generate large-scale synthetic identity document datasets programmatically using different parameters with low time and monetary costs. In the future, we will integrate the pipeline with mobile device capturing to produce identity documents using different devices, backgrounds, and lighting conditions.

\begin{table}
\small
\centering
\caption{Breakdown of Time and Monetary Costs of the Proposed IDNet Pipeline.}
\label{tab:costs}
\begin{tabular}{ |p{1.5cm}|p{1.4cm}|p{1.4cm}|p{1.6cm}|p{1.5cm}|p{1cm}|p{1.3cm}|p{1.5cm}| } 
 \hline
  & Template & Metadata & Portrait preprocess &Parameter tuning& Filling & Total & Avg (per doc)\\ \hline \hline
Time cost (seconds)& 231 & 555 &13,153&\textcolor{black}{4,040}&\textcolor{black}{103,249}&\textbf{\textcolor{black}{121, 228}}&\textbf{0.14}\\ \hline
Monetary cost (\$) & 0.12 & 0.06 &7.3&\textcolor{black}{2.2}&\textcolor{black}{59.5}&\textbf{\textcolor{black}{69.2}}&\textbf{0.00008}\\ \hline
\end{tabular}
\end{table}

\section{IDNet Quality Evaluation}
\label{sec:quality}

The objective of this research is not to provide a set of documents that exactly resemble corresponding real-world identity documents. Instead, we attempt to provide a comprehensive and diverse set of meaningful documents that follow most identity document design standards and meet the research requirements for privacy-preserving document analysis and fraud detection~\cite{el2020practical}\cite{abedjan2015profiling}. Despite the unavailability of real identity document datasets, we compared IDNet to existing benchmarks such as MIDV and STDID on metadata quality, document fidelity, fraud stealthiness, and task utility, as detailed in the following sections.

\subsection{Metadata Quality} \label{Metadata Diversity}
\label{sec:diversity}

\begin{wrapfigure}{r}{0.6\textwidth}
\vspace{-4mm}
\begin{center}
\centerline{\includegraphics[width=0.6\columnwidth]{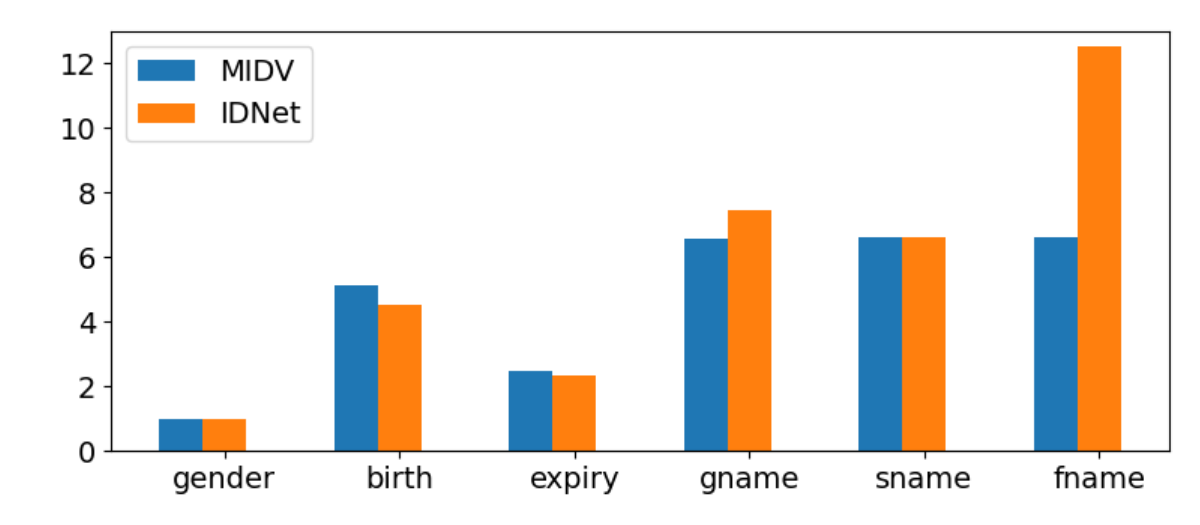}}
\caption{Column entropy comparison for Slovak ID documents. 
}
\label{fig:entropy}
\end{center}
\vspace{-10mm}
\end{wrapfigure}

In this section, we used the MIDV-2020 benchmark, which also contains identity documents from the ten European countries as listed in Tab.~\ref{tab:type-statistics} as a baseline to show that the generated metadata information to be filled into text fields (e.g., name, gender, DOB, address, date of issue, date of expiration, and so on) in IDNet are of similar or higher quality. This study measured the text fields in three aspects: uniqueness of text fields such as document ID number and personal ID number; diversity of text fields such as gender, birth date, and expiry date; cross-field dependency, for example, the birth date must be before the issue date.

\begin{enumerate}
\item Uniqueness: In IDNet, every document ID number is unique for each country. In MIDV-2020, however, there exists duplicate ID numbers, e.g., "668174749" occurred twice in the Finnish ID documents. 
\item Diversity:  We used entropy to measure diversity. Larger entropy indicates more diversity in data~\cite{abedjan2015profiling}~\cite{chopra2023cowrangler}. We measured the entropy of gender, surname, given name, birth year, issue year, and expiry year. These are the text fields shared across the ten countries. {For national-level documents, such as passports, fields such as nationality and {issuing} authority are usually the same within the same country, so they were excluded in the diversity study.} Fig.~\ref{fig:entropy} showed the comparison results for one European country, Slovak, which indicates that IDNet achieved similar or higher entropy than MIDV-2020. The results for other European countries are similar, which can be found in the Appendix~\ref{sec:entropy}. 
\item Dependency: Both MIDV-2020 and IDNet satisfy the dependency constraints such as that the birth date should be before the issue date; the expiry date should be set according to a fixed valid period, and so on. 
\end{enumerate}

\eat{
\begin{figure}[!htbp]
\begin{center}
  \includegraphics[width=0.7\linewidth]{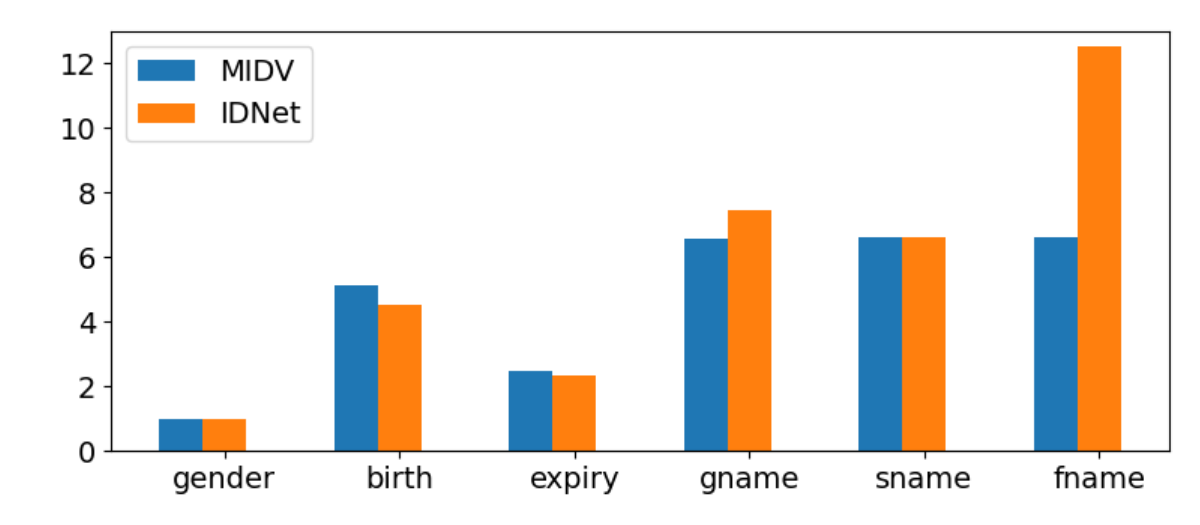}
\end{center}
\vspace{-2mm}
\caption{Comparison of generated text field entropy between MIDV and IDNet for Slovak ID documents. "birth" stands for the birth year, "expiry" stands for the expiry year, "gname" stands for the given name, "sname" stands for the surname, "fname" stands for the full name.
}
\label{fig:text_field_entropy}
\end{figure}
}

\subsection{Document Fidelity}
\label{sec:fidelity}
Several factors may affect the fidelity of the IDNet document and cause differences between IDNet documents and their real-world counterparts. First, the generative models may add {noise} when generating the ID templates, such as distortion in the background. Second, inconsistent font style, size, and spacing were used when filling the metadata information into the template. To evaluate the fidelity of IDNet documents, we adopt a standard and widely used metric, Structural Similarity Index (SSIM)~\cite{hore2010image}.

\begin{figure}[!htbp]

\begin{center}
  \includegraphics[width=\linewidth]{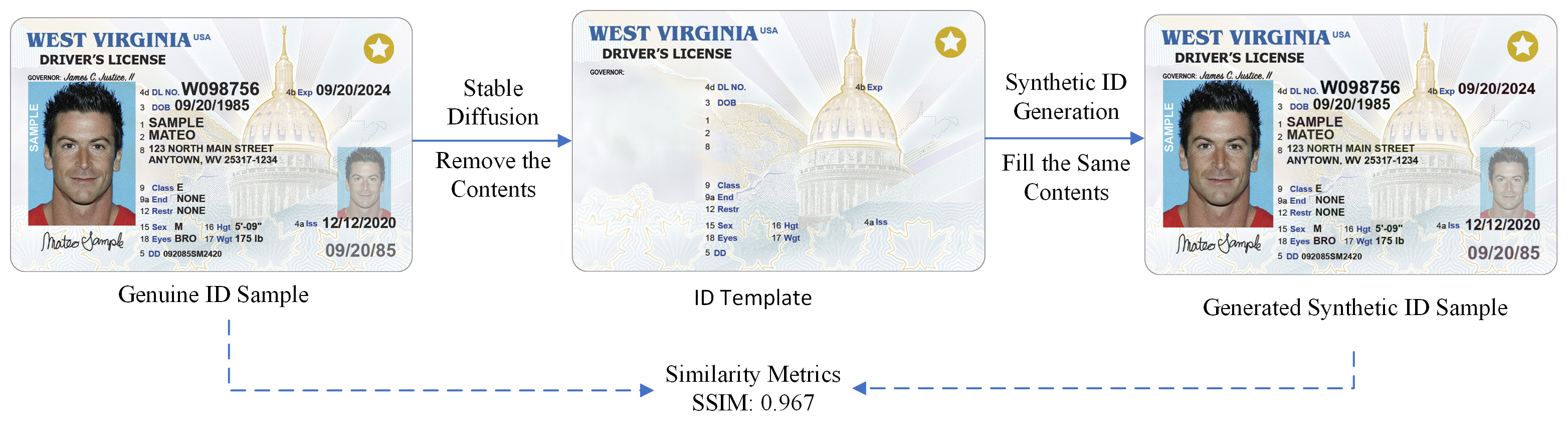}
\end{center}
\vspace{-5mm}
\caption{Illustration of the identity document template fidelity evaluation process.}
\label{fig:template_evaluation}
\end{figure}

SSIM can indicate the perceived change in structural information, brightness, and contrast. Evaluating the fidelity of generated images typically requires ground-truth images, i.e., the genuine ID templates in our case. We used the official ID samples {from publicly available government websites} as ground truth. We filled our generated blank ID templates with the same contents of the genuine sample IDs. Then, we calculated the SSIM between the authentic sample IDs and the generated synthetic sample IDs to evaluate the IDNet’s fidelity quantitatively. The evaluation pipeline is illustrated in Figure~\ref{fig:template_evaluation}.

As a result,
the average SSIM of 20 ID document templates is \textcolor{black}{$0.888$}, with a standard deviation of \textcolor{black}{$0.0058$}. 
SSIM is a metric closely aligned with human perception, evaluating changes in structure, brightness, and contrast~\cite{setiadi2021psnr}. It ranges from $-1$ to $1$, with $1$ indicating the same structural pattern between two images.  
Our SSIM result of $0.853$ indicates the generated templates are of satisfactory visual fidelity, and meet our fidelity goal.

\subsection{Stealthiness of the Generated Fraud Data }
\label{sec:stealthiness}

We also evaluated the stealthiness of the fraud identity document samples generated in IDNet. 
We compare these samples with another synthetic fraud ID dataset named SIDTD~\cite{boned2024synthetic}. We chose the Albanian ID, which is used on both IDNet and SIDTD, for evaluation. Specifically, we calculated the SSIM between each fraud ID image and the corresponding ID image without fraud. 
The detailed results are shown in Tab.~\ref{tab:img_qual_comparison}. We observed that all types of fraud document samples in IDNet obtained higher SSIM than the SIDTD dataset, indicating that IDNet is made up of more stealthy fraud samples, which are more similar to the corresponding samples without fraud compared to SIDTD.


\vspace{-10pt}
\begin{table}[h]
    \centering
    \caption{Image quality (SSIM) comparison between IDNet and SIDTD ~\cite{boned2024synthetic}.}
    \resizebox{\textwidth}{!}{
    \begin{tabular}{ccccc|c}
    \toprule
     \multicolumn{5}{c}{Ours} & \multicolumn{1}{|c}{Baselines}\\
     
      & Face Morphing & Portrait Substitution & Mixed Fraud Face & Text Replacement & SIDTD \\
    \midrule
    & 0.97 & 0.99 & 0.99 & 0.94 & 0.92\\
    \bottomrule
    \end{tabular}
    }
    \label{tab:img_qual_comparison}
\end{table}

\subsection{Utility of the Synthetic Portrait Photo Dataset used in IDNet}
\label{sec:morphing-utility}
\begin{table}[!bp]
\centering
\caption{Real-world evaluation results of face morphing detection models trained on IDNet.}
\resizebox{\textwidth}{!}{
\begin{tabular}{cc|c|c|c|c}
\toprule
\multirow{2}{*}{Training Dataset} & \multirow{2}{*}{Models} & \multicolumn{4}{c}{Test Datasets} \\ \cmidrule{3-6} 
                                     &                         & IDNet (synthetic)   & FRLL (real-world)            & FERET (real-world) & FRGC (real-world)     \\ \midrule
\multirow{4}{*}{IDNet (synthetic)}   & SPL-MAD                 & 92.8                & 87.3            & 74.8   & 77.3      \\
                                     & PW-MAD                  & 97.6                & 89.7            & 78.0   & 80.5      \\
                                     & Inception               & 98.2                & 90.7            & 78.4   & 81.6      \\
                                     & MixFaceNet-S            & 98.7                & 92.8            & 79.1   & 82.0     \\ \midrule
\multirow{4}{*}{FRLL (real-world)}   & SPL-MAD                 & 84.1                & 90.5            & 74.2   & 77.5      \\
                                     & PW-MAD                  & 87.3                & 97.8            & 76.0   & 80.0      \\
                                     & Inception               & 89.5                & 96.8            & 77.9   & 81.2      \\
                                     & MixFaceNet-S            & 89.7                & 98.3            & 78.4   & 81.3     \\ \midrule
\multirow{4}{*}{FERET (real-world)}  & SPL-MAD                 & 87.5                & 87.5            & 80.9   & 77.9      \\
                                     & PW-MAD                  & 92.3                & 93.2            & 84.3   & 82.3      \\
                                     & Inception               & 93.5                & 94.3            & 84.5   & 83.0      \\
                                     & MixFaceNet-S            & 94.0                & 94.6            & 85.2   & 83.9     \\ \midrule
\multirow{4}{*}{FRGC (real-world)}   & SPL-MAD                 & 86.9                & 88.9            & 75.1   & 82.1      \\
                                     & PW-MAD                  & 91.9                & 92.8            & 79.2   & 85.6      \\
                                     & Inception               & 92.5                & 93.5            & 79.8   & 85.8      \\
                                     & MixFaceNet-S            & 92.8                & 93.3            & 80.5   & 87.2     \\ \bottomrule
\end{tabular}
}
\label{tab:face_morphing_test_on_real}
\end{table}
IDNet used a publicly available synthetic portrait photo dataset~\cite{generated.photos}. We evaluated its quality using the utility of the face morphing task.
Task utility is an important metric in evaluating the quality of synthetic datasets~\cite{jordon2022synthetic} by comparing the task accuracy on the synthetic datasets and the real-world datasets. 
Leveraging the availability of real-world face morphing data, such as FERET~\cite{sarkar2020vulnerability}, FRGC~\cite{frgc}, and FRLL~\cite{frll}, we evaluated the utility of the artificially generated portrait photos dataset~\cite{generated.photos} used in our IDNet dataset from two perspectives, which are (1) training models on synthetic data and then assessing their performance on real-world data, and (2) analyzing whether models maintain their relative performance rankings when trained on both synthetic and real-world data~\cite{jordon2022synthetic}. 

In the first perspective, we evaluate how models trained on IDNet transfer to real-world datasets in terms of face morphing detection accuracy. This study assumes that models trained on synthetic datasets will be directly applied to real-world applications. For comparison, we also test how models trained on real-world datasets transfer to other real-world datasets. In the second perspective, we study whether the rankings of a set of models would be the same on synthetic or real-world data in terms of face morphing detection accuracy following~\cite{jordon2022synthetic}. This study assumes that the synthetic data is used to select the best model that will be later trained on a real-world dataset for better performance.

In our evaluation, we used three real-world face morphing detection datasets~\cite{sarkar2022gan}, FRLL~\cite{frll},  FERET~\cite{sarkar2020vulnerability}, and FRGC~\cite{frgc}. To compose our face morphing dataset, for each photo from the IDNet's synthetic portrait photo dataset (i.e., the $5,979$ synthetic portrait photos as mentioned in Sec.~\ref{sec:generation}), we applied FaceMorpher~\cite{sarkar2022gan} to morph it with another photo randomly selected from the same dataset using a morph percentage of $0.5$. We split each face morphing dataset into a training set containing $80\%$ of the data and a test set containing $20\%$ of the data. All the facial images are cropped using the MTCNN~\cite{zhang2016joint} with an extension rate of $5\%$ to include the whole face following~\cite{damer2021pw}. We implemented four face morphing detection methods, PW-MAD~\cite{damer2021pw},  Inception~\cite{ramachandra2019detecting}, MixFaceNet-S~\cite{boutros2021mixfacenets}, and SPL-MAD~\cite{fang2022unsupervised}. The evaluation results on real-world datasets FRLL,  FERET, and FRGC, using face morphing detection models trained on IDNet's face morphing dataset, are shown in Table~\ref{tab:face_morphing_test_on_real}. It illustrates that despite the performance drops due to differences in background, brightness, etc, the models trained on IDNet maintained acceptable accuracy on the real-world datasets compared with models trained on real-world datasets.

In addition, models trained on IDNet achieved similar domain-transfer performance as models trained on some real-world datasets. 

On the other hand, the relative performance comparisons of four face morphing detection methods are shown in Table~\ref{tab:face_morphing_ranking}. The results indicate that the rankings remain consistent across all four datasets, with the sole exception being the performance of PW-MAD and Inception on the FRLL dataset. In other words, the synthetic portrait photo dataset is a good option for replacing real-world datasets to study the relative performance of different face morphing detection methods. 

\begin{table}[!htbp]
\centering
\caption{Relative performance rankings comparison of different face morphing detection methods on different datasets.}
\resizebox{\textwidth}{!}{
\begin{tabular}{c|c|c|c|c|c|c|c|c} 
\hline
\multirow{2}{*}{Models} & \multicolumn{2}{c|}{IDNet (synthetic)} & \multicolumn{2}{c|}{FRLL (real-world)} & \multicolumn{2}{c|}{FERET (real-world)} & \multicolumn{2}{c}{FRGC (real-world)}  \\ 
\cline{2-9}
                        & Accuracy & Rank                        & Accuracy & Rank                        & Accuracy & Rank                         & Accuracy & Rank                        \\ 
\hline
SPL-MAD                 & 92.8     & 4                           & 90.5     & 4                           & 80.9     & 4                            & 82.1     & 4                           \\
PW-MAD                  & 97.6     & 3                           & 97.8     & 2                           & 84.3     & 3                            & 85.6     & 3                           \\
Inception               & 98.2     & 2                           & 96.8     & 3                           & 84.5     & 2                            & 85.8     & 2                           \\
MixFaceNet-S            & 98.7     & 1                           & 98.3     & 1                           & 85.2     & 1                            & 87.2     & 1                           \\
\hline
\end{tabular}
}
\label{tab:face_morphing_ranking}
\end{table}

In addition, since all the portrait photos used in our dataset are artificially generated, the usage of our data can address the privacy concerns of using real-world datasets~\cite{sarkar2020vulnerability}\cite{damer2022privacy} in evaluating face morphing detection methods.

{\color{black}
\subsection{Utility of the Text-field Fraud Replacement Dataset in IDNet}
\label{sec:text-replacement-utility}
\begin{table}[h]
\scriptsize
\centering
\caption{\small Utility (detection accuracy) of the models trained on IDNet and SIDTD for detecting inpaint-and-rewrite and copy-and-move frauds for Finland ID cards.}
\begin{tabular}{cc|c|c}

\toprule
\multirow{2}{*}{Training Dataset} & \multirow{2}{*}{Models} & \multicolumn{2}{c}{Test Datasets} \\ \cmidrule{3-4} 
                                     &                         & IDNet-Finland   & SIDTD-Finland                \\ \midrule
\multirow{3}{*}{IDNet-Finland}   & EfficientNet               & 100.0                & 100.0               \\
                                     & ResNet18                 & 100.0                & 100.0            \\
                                     & ResNet50              & 100.0                & 100.0                \\ \midrule
\multirow{3}{*}{SIDTD-Finland}   & EfficientNet               & 67.2                & 100.0           \\
                                     & ResNet18                 & 67.2                & 100.0           \\
                                     & ResNet50              & 67.2                & 100.0           \\ \bottomrule
\end{tabular}
\label{tab:finland_text_field_replacement}
\end{table}

\begin{table}[h]
\scriptsize
\centering
\caption{\small Utility (detection accuracy) of the models trained on IDNet and SIDTD for detecting inpaint-and-rewrite and copy-and-move frauds for Greece passports.}
\begin{tabular}{cc|c|c}
\toprule
\multirow{2}{*}{Training Dataset} & \multirow{2}{*}{Models} & \multicolumn{2}{c}{Test Datasets} \\ \cmidrule{3-4} 
                                     &                         & IDNet-Greece   & SIDTD-Greece                \\ \midrule
\multirow{3}{*}{IDNet-Greece}   & EfficientNet              & 100.0                & 100.0                \\
                                     & ResNet18                 & 100.0                & 96.8            \\
                                     & ResNet50              & 100.0                & 99.1               \\ \midrule
\multirow{3}{*}{SIDTD-Greece}   & EfficientNet               & 67.7                & 100.0            \\
                                     & ResNet18                & 67.7                & 100.0          \\
                                     & ResNet50             & 67.7                & 100.0            \\ \bottomrule
\end{tabular}
\label{tab:greece_text_field_replacement}
\end{table}

\noindent
\textbf{Detection of Inpaint-and-Rewrite, and Crop-and-Move.}
To evaluate the synthetic fraud data generated from the text-field fraud replacement pattern, we focus on the inpaint-and-rewrite pattern and the crop-and-replace pattern, since they were also used in SIDTD \cite{boned2024synthetic} and the utility of the fraud detection models trained on IDNet can be evaluated on SIDTD. We selected two European identification documents, the Finland ID and the Greece passport, for evaluation, as both document templates have been used in IDNet and SIDTD \cite{boned2024synthetic}. (The results on identity documents from other European countries shared by the two datasets are similar.) For each fraud pattern and each document type, our IDNet dataset contains $5979$ positive samples (w/ fraud) and $5979$ negative samples (w/o fraud). However, SIDTD only contains $100$ negative samples for each of $10$ types, and $1078$ and $144$ positive samples of the inpaint-and-rewrite and the crop-and-replace patterns for all $10$ types in total, respectively. Particularly, there are $112$ negative samples of the inpaint and rewrite fraud and $10$ samples of the copy-and-move fraud for the Finland ID card, and $109$ negative samples of the inpaint and rewrite fraud and $12$ samples of the copy-and-move fraud, for the Greece passport.

Table \ref{tab:finland_text_field_replacement} and Table \ref{tab:greece_text_field_replacement} illustrate the results for Finland IDs and the Greece passports, respectively. From these results, we can interpret that when the model is trained on the datasets from IDNet and tested against the corresponding datasets from SIDTD, the model successfully captures the fraud patterns of SIDTD. However, in the reverse scenario, where the model is trained on SIDTD fraud patterns and tested on the datasets of IDNet, the model fails to capture the underlying fraud patterns.

\section{Use Cases}
\label{sec:cases}

The IDNet dataset is poised to significantly enhance a wide array of applications, spanning from privacy-centric fraud identification to the nuanced detection of face morphing, cross-type analysis, and the alignment and unification of schemas through Large Language Models (LLMs). The forthcoming sections provide a detailed exploration of these application scenarios, illustrating the dataset's versatility and its potential to catalyze advancements across diverse domains in identity document management.


\subsection{Privacy-Preserving Fraud Detection}
\label{sec:pp-fraud}
\begin{table}[]
    \centering
        \vspace{-10pt}
    \caption{Fraud ID detection on West Virginia Driver License dataset.}
    \begin{tabularx}{\textwidth}{c *{4}{>{\centering\arraybackslash}X}}
    \toprule
     & \multicolumn{4}{c}{Privacy-Preserving Method}\\
    \cmidrule(lr){2-5}
     Task & None & Masking & PixelDP (L=0.1) & PixelDP (L=1.0) \\
    \midrule
     Face Morphing & 98.65 & 50.00 & 90.30 & 52.55 \\
     portrait substitution & 88.95 & 50.00 & 91.05 & 87.25 \\
     Mixed Fraud Face & 89.30 & 50.00 & 86.25 & 68.80 \\
     Text Replacement & 100.00 & 50.00 & 98.99 & 50.00 \\
    \bottomrule
    \end{tabularx}
    \vspace{-10pt}
    \label{tab:wv_privacy_preserving}
\end{table}

\begin{table}[]
    \centering
            \vspace{-10pt}
        \caption{Fraud ID detection on Arizona Driver License dataset.}
    \begin{tabularx}{\textwidth}{c *{4}{>{\centering\arraybackslash}X}}
    \toprule
     & \multicolumn{4}{c}{Privacy-Preserving Method}\\
    \cmidrule(lr){2-5}
     Task & None & Masking & PixelDP (L=0.1) & PixelDP (L=1.0) \\
    \midrule
     Face Morphing & 98.10 & 50.00 & 88.30 & 52.65 \\
     Portrait substitution & 88.75 & 50.00 & 91.30 & 89.05 \\
     Mixed Fraud Face & 88.80 & 50.00 & 84.25 & 70.35 \\
     Text Replacement & 100.00 & 50.00 & 100.00 & 50.00 \\
    \bottomrule
    \end{tabularx}
            \vspace{-10pt}
    \label{tab:az_privacy_preserving}
\end{table}

In releasing our novel dataset for identity document analysis and fraud detection, it is important to consider the privacy concerns. The dataset's intrinsic value lies not only in its comprehensive coverage and potential to revolutionize fraud detection but also in the sensitive nature of the information it encompasses (e.g., the photo identity, name, and so on). To responsibly harness this value while safeguarding individual privacy, the design and deployment of a privacy-preserving algorithm is not just beneficial but imperative. Thus, in this section, our goal is to evaluate the performance of our dataset when it meets with the current privacy-preserving algorithm.

To achieve this goal, we selected two standard privacy-preserving algorithms: Masking~\cite{masking2}~\cite{ren2018learning} and PixelDP \cite{lecuyer2019certified}. The algorithm details and settings are described as follows:  
\begin{itemize}
    \item \textbf{Masking. } As shown in Figure~\ref{fig:wv_examples} (b) and \ref{fig:az_examples} (b), we applied masks to those regions that contain sensitive information, \emph{i.e.} zeroing the pixel values. In this way, all the representations were completely erased without any possibility of being recognized. While this approach, which is also termed redaction,  has been widely used~\cite{masking2}~\cite{ren2018learning}~\cite{zhou2023privacy}, it leads to information loss and may disable analysis tasks that rely on the redacted information.
    \item \textbf{PixelDP. } Introduced by \cite{lecuyer2019certified}, PixelDP is a robust privacy-preserving method via adding noise sampled from a specific distribution to the input or intermediate features. In this paper, we focused on the input-level PixelDP, directly applying Gaussian noise to the input examples. Specifically, We set $\epsilon=1.0$ and $\delta=0.05$, which provides $(1.0, 0.05)$-DP guarantee. The perturbations are sampled from a Gaussian distribution with zero mean and standard deviation $\sigma=\sqrt{2\ln(\frac{1.25}{\delta})\Delta_{p,2}L/\epsilon}$, where we use $L_2$-norm (\emph{i.e.} $p=2$) and $\Delta_{2,2}=1$. With larger $L$, the perturbations will be more significant, and thus distort more information.    
\end{itemize}

\paragraph{Experimental Settings. } We evaluated these two algorithms on four different fraud ID detection tasks, including face morphing detection, portrait substitution detection, mixed fraud face detection, and text replacement detection. For the first three tasks in the face regions, we used a MixFaceNet-S \cite{tan2019mixconv} model as the detector, with $embedding\_size=128$, $width\_scale=1.0$, $gdw\_size=1024$, and shuffling disabled. For text replacement detection, we used a simple convolutional network with $5$ convolutional layers and $2$ fully-connected layers. All the models were trained for $200$ epochs with $lr=0.1$. The validation dataset consists of $1000$ positive and negative examples.

\paragraph{Results.}
The results are shown in Table \ref{tab:wv_privacy_preserving} and Table \ref{tab:az_privacy_preserving}, and the example images are illustrated in Figure \ref{fig:wv_examples} and Figure \ref{fig:az_examples}. Directly masking almost disabled the fraud detection capability, although it completely preserved the privacy of the sensitive information. That's because the helpful representations for fraud detection were also completely erased by masking, and the detector learned nothing from the training examples. 
Meanwhile, PixelDP with smaller $L$ barely degraded the fraud detection performance (and even slightly improved it, serving as a kind of data augmentation), while the sensitive information were still highly recognizable. With larger $L$, the sensitive information werebetter concealed, at the cost of a nonnegligible fraud detection performance drop, since the helpful representations for fraud detection were also distorted (except for portrait substitution detection, which is a relatively easier task.) 

\paragraph{Analysis.} From the results of two baseline privacy-preserving algorithms, Masking and PixelDP, we observed a significant gap in their capabilities in balancing utility and privacy. They cannot achieve satisfactory fraud detection performance while protecting sensitive information from being recognized. How to simultaneously conceal sensitive information and keep helpful representations for fraud detection is yet an unsolved conundrum. Our released dataset can serve as a new benchmark, bringing new challenges for privacy-preserving algorithms. 

\begin{figure}[h]
  \centering
  \subfigure[Original]{
      \includegraphics[width=0.4\textwidth]{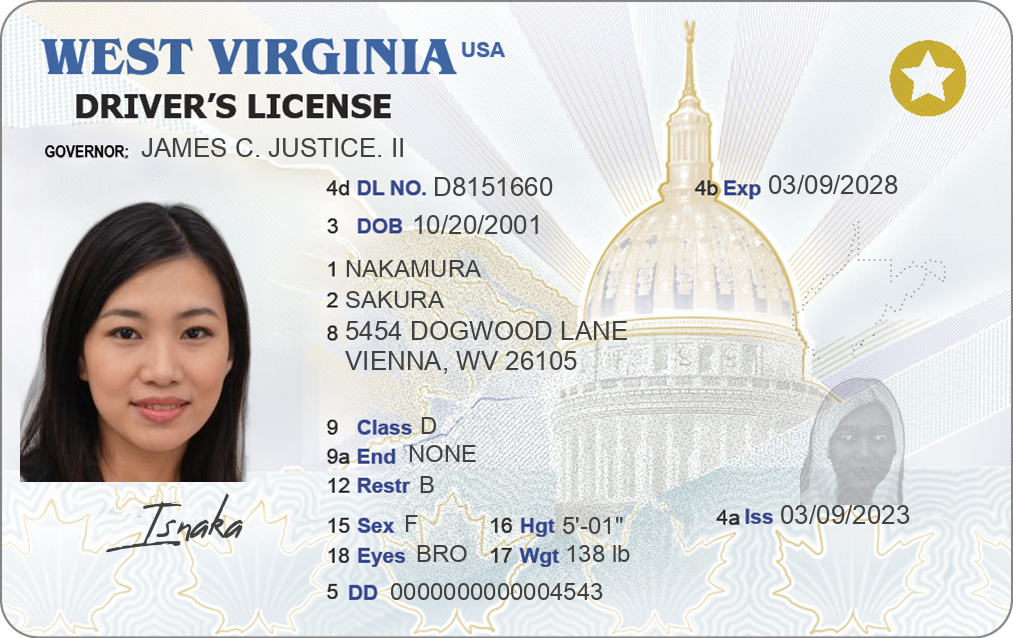}
  }
  \subfigure[Masking]{
      \includegraphics[width=0.4\textwidth]{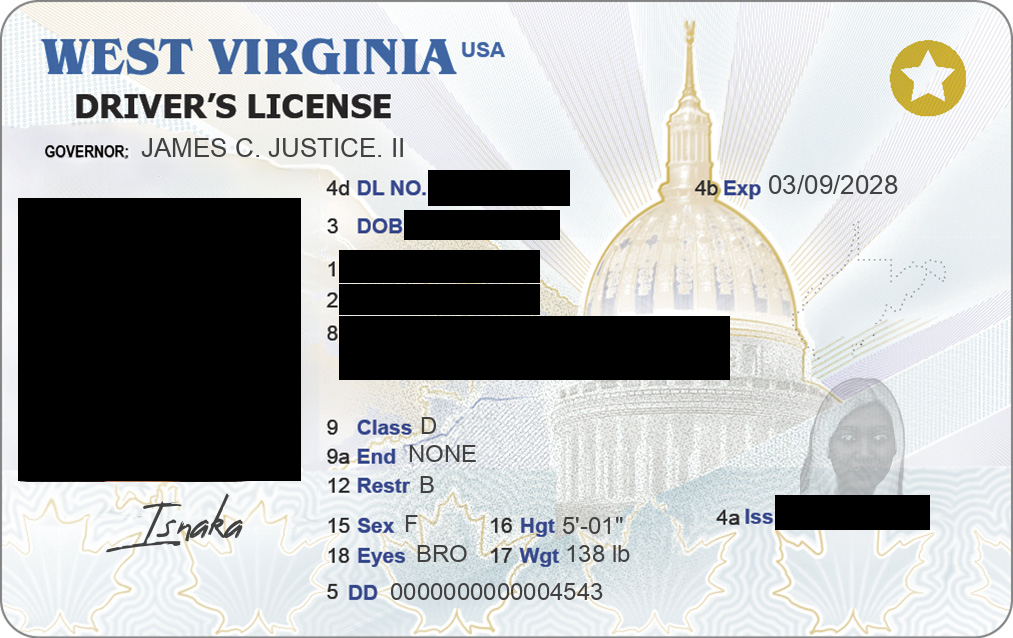}
  }
  \subfigure[PixelDP ($L=0.1$)]{
      \includegraphics[width=0.4\textwidth]{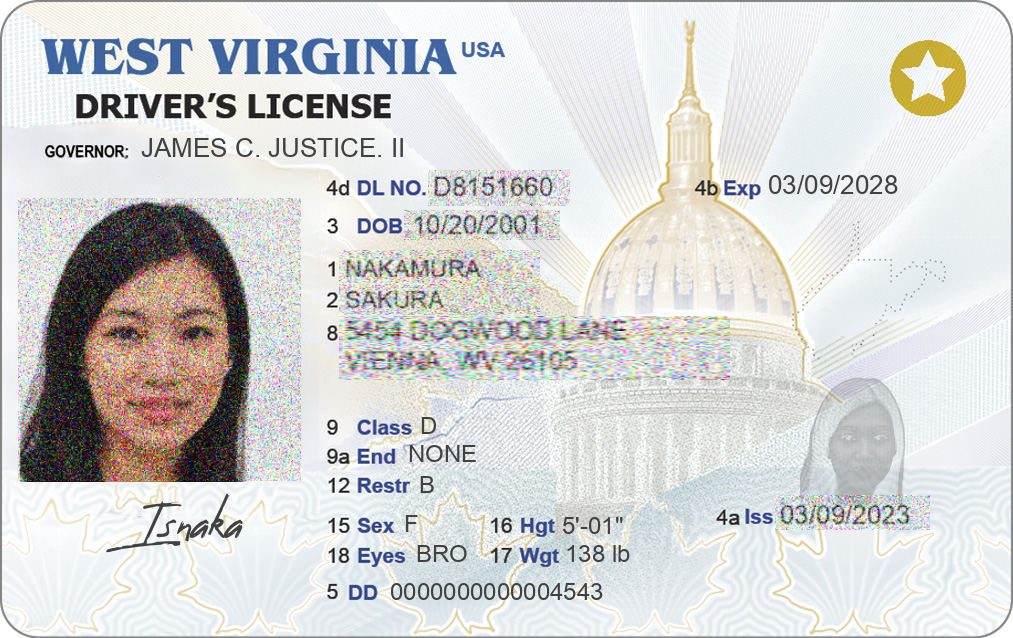}
  }
  \subfigure[PixelDP ($L=1.0$)]{
      \includegraphics[width=0.4\textwidth]{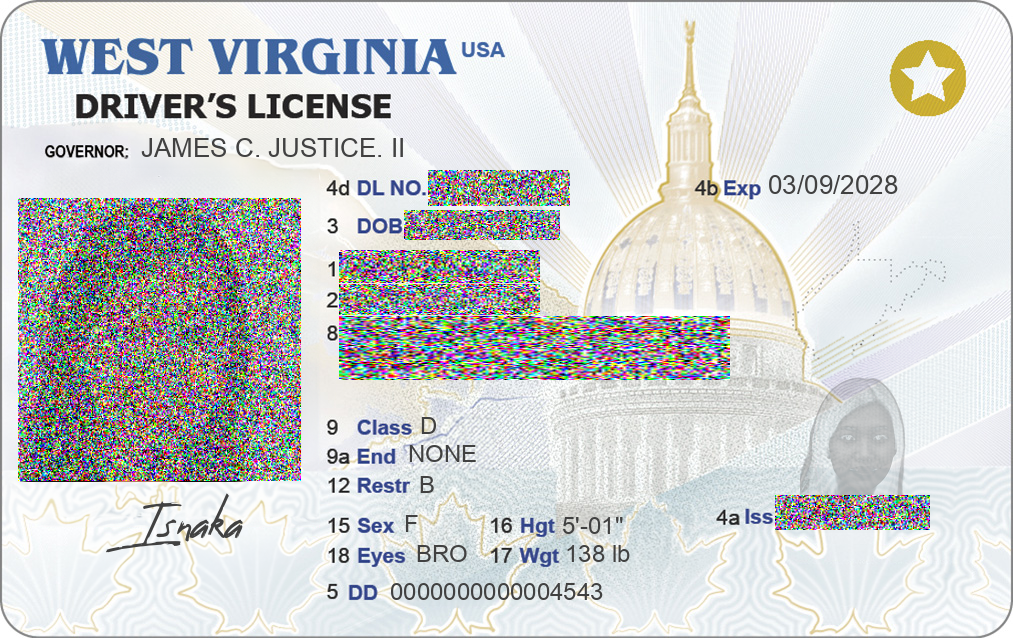}
  }
  \caption{West Virginia Driver License examples.}
  \label{fig:wv_examples}
\end{figure}

\begin{figure}[h]
  \centering
  \subfigure[Original] 
  {
      \includegraphics[width=0.4\textwidth]{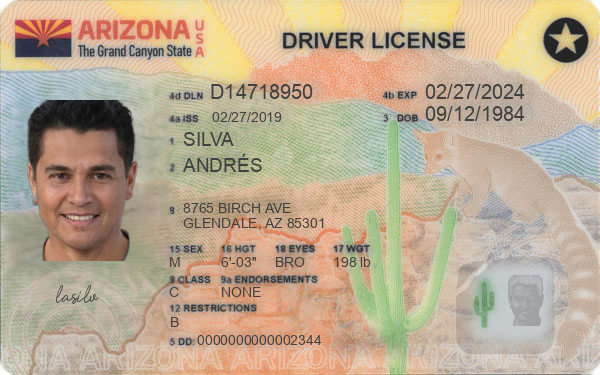}
  }
  \subfigure[Masking]
  {
      \includegraphics[width=0.4\textwidth]{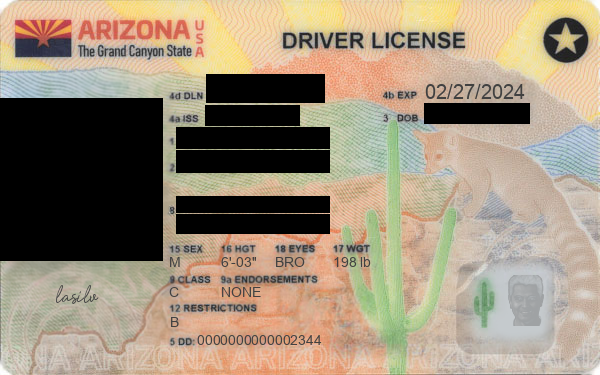}
  }
  \subfigure[PixelDP ($L=0.1$)]
  {
      \includegraphics[width=0.4\textwidth]{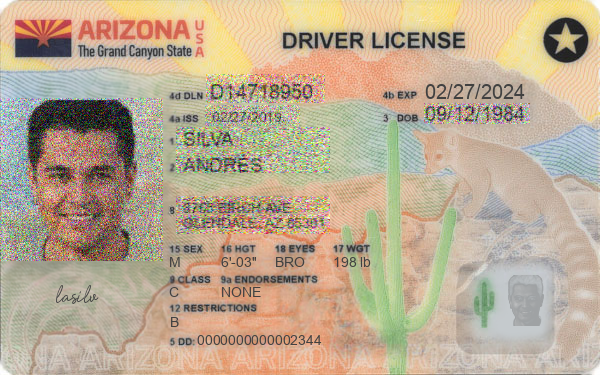}
  }
  \subfigure[PixelDP ($L=1.0$)]
  {
      \includegraphics[width=0.4\textwidth]{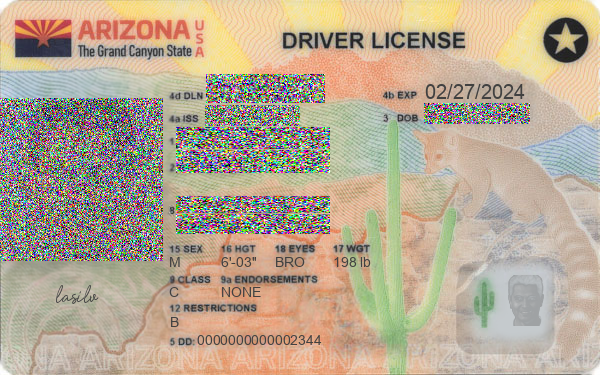}
  }
  \caption{Arizona Driver License examples.}
  \label{fig:az_examples}
\end{figure}

\subsection{Face Morphing Detection}
\label{sec:morphing}

Face morphing poses a threat to the security and reliability of face recognition systems, and thus face morphing detection becomes a focus in enhancing these systems, ensuring they remain robust against fraudulent attempts to bypass or deceive identity {proofing} processes.
In Sec.~\ref{sec:morphing-utility}, we discussed the utility of the face morphing task on the synthetic portrait photo dataset used in IDNet. 

However, once the portrait photo is added to each type of identity document template, the photo's background, color, transparency, size, and cropping will change among different templates.
In this section, we further study how face morphing detection methods work in identifying IDs with morphed faces in IDNet's identity documents. 

\eat{
\begin{figure}[!htbp]
    \subfigure[Genuine Face 1]{
        \includegraphics[width=0.28\columnwidth]{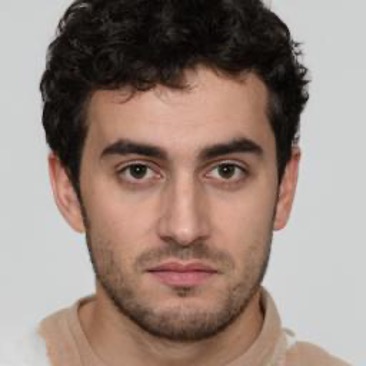}
    }
    \hspace{0.2cm}
     \subfigure[Morphed Face]{
        \includegraphics[width=0.28\columnwidth]{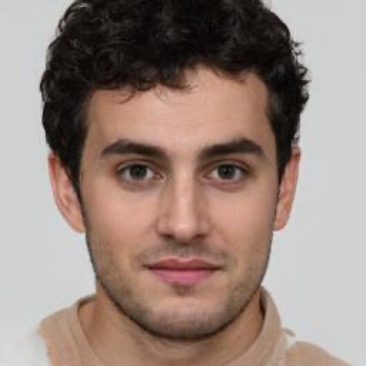}
    }
    \hspace{0.2cm}
    \subfigure[Genuine Face 2]{
        \includegraphics[width=0.28\columnwidth]{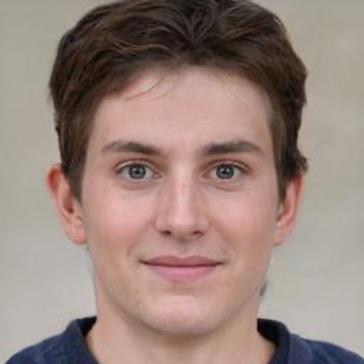}
    }

    \caption{Sample of genuine faces and morphed faces used in our dataset. The morphed face in (b) is generated by morphing genuine faces in (a) and (c).}
    
     \label{fig:face_morphing}
\end{figure} 
}

In our experiments, we adopted face morphing detection methods~\cite{ngan2020face} that differentiate between the two classes, morphed portrait photos and normally generated portrait photos. An example of normally generated portrait photos and morphed portrait photos used IDNet can be found in Fig.~\ref{fig:portrait-morphing-fraud}. For each ID type, we split all the portrait photos with the face morphing fraud applied into a training set and a test set. The training set contains the first $4,979$ normal portrait photos and their corresponding $4,979$ morphed portrait photos. The test set contains the rest $1,000$ portrait photos without frauds and their corresponding $1,000$ morphed portrait photos. We build a simple binary classification network following~\cite{damer2022privacy} as our face morphing detector. The face recognition backbone MixFaceNet-S~\cite{boutros2021mixfacenets} is adopted as the representation learning backbone. A fully-connected layer is connected to the representation learning backbone and serves as a classification head. We have also evaluated two other representation learning backbones, including SPL-MAD~\cite{fang2022unsupervised}, PW-MAD~\cite{damer2021pw} and Inception~\cite{ramachandra2019detecting} as baselines.

\paragraph{Experimental Settings.} In our experiment, we used binary cross-entropy loss to supervise the training. We used Adam optimizer with a learning rate of $0.1$ and a weight decay of $10^{-5}$ to optimize the weights in the networks. All the models are trained for $200$ epochs. In addition, an early stopping strategy with the patience of $20$ epochs is used to avoid over-fitting in our training. The input of the network is reshaped to $256\times 256\times 3$. 

\begin{table}[!htbp]
    \centering
    \caption{Face morphing detection results.}
    \begin{tabularx}{1\textwidth}{c *{4}{>{\centering\arraybackslash}X}}
    \toprule
     & \multicolumn{4}{c}{Morphing Detection Accuracy (\%)}\\
    \cmidrule(lr){2-5}
     Datasets & SPL-MAD& PW-MAD  & Inception  & MixFaceNet-S  \\
    \midrule
     Arizona Driver's License (US) & 91.9 & 97.3 & 97.6 & 98.1\\
     West Virginia Driver's License (US) & 92.8 & 97.6 & 98.2 & 98.7\\
     Nevada ID (US)& 94.2 & 98.9 & 99.2 & 99.5\\
     Greece Passport & 98.7 & 100.0 & 99.9 & 100.0\\
     Finland ID & 98.5 & 99.9& 99.7 & 100.0\\
     Spain ID & 98.9 & 99.8& 99.9 & 99.9\\
    \bottomrule
    \end{tabularx}
    \vspace{-10pt}
    \label{tab:face_morphing}
\end{table}

\paragraph{Experimental Results.} 
We evaluated all face morphing detection methods on 6 different types of ID documents in IDNet, including Arizona Driver's License (US), West Virginia Driver's License (US), Nevada ID (US), Greece Passport, Finland ID, and Spain ID. 
The findings, detailed in Table~\ref{tab:face_morphing}, reveal that US IDs, with their more intricate designs, pose greater challenges for face morphing detection methods compared to their European counterparts. Despite these variations, the comparative effectiveness of the different detection methods remained consistent across all ID types, hinting at the universal applicability of different types of ID documents for evaluating and comparing different face morphing detection methods
Notably, MixFaceNet-S emerged as the most effective one among the methods we evaluated.

\subsection{Cross-Type Analysis}
\label{sec:cross-domain}

As fraudsters refine their techniques and extend their operations beyond traditional boundaries, the need to study and understand these methods across various document types and states becomes imperative. Our IDNet benchmark consists of 20 different types, so it can be used to study the model performance on {multiple types}, e.g., how models trained on one type of identity document can generalize to other types. In this way, it will be better to understand the importance of generating diverse types of data.
To showcase such analysis, we applied fraud-detection models trained on the West Virginia (WV) driver’s license documents to detect the fraud driver’s license documents issued by the Arizona (AZ) state, and vice versa. As shown in Tab. 10, the accuracy of the face morphing detection (using the MixFaceNet-S algorithm) drops from 98.65\% to 50.55\% while the model trained on the WV domain is used in the AZ domain, and drops from 98.1\% to 54.1\% when the model trained on the AZ {type} is applied to the WV {type}. We observed similar accuracy drops for the text replacement fraud detection models. The results showed that the fraud detection models heavily rely on {type}-specific features such as background patterns, transparency, size, etc. While the Portrait substitution fraud detection model is relatively more robust to {type}-specific changes (i.e., portrait substitution detection more relies on face features such as age, head position, etc.), significant accuracy drops (e.g., from 88.95\% to 71.20\% and from 88.75\% to 80.05\%) have also been observed. We conducted more cross-{type} analysis. For example, in Tab.~\ref{tab:cross_domain_val-id}, we trained models using Finland ID card documents and tested the models using Spain ID card documents. Similarly, in Tab.~\ref{tab:cross_domain_val-passport}, we trained models using Greece passport documents and tested the models using Albania passport and Greece passport. In Tab.~\ref{tab:cross_domain_val-az-file-types}, the models were trained on the AZ driver’s license documents, but tested on Spain ID card documents and Albania passport documents. In Tab.~\ref{tab:cross_domain_val-finland-file-types}, the models were trained on the Finland ID card documents, while tested using Albania passport documents and AZ Driver’s license.
These results {indicate} a performance gap {exists} when applying the fraud detector to {a new type that is unseen during the training process}. Such results showed the importance of our work, which aims to generate diverse types of data. We also hope our study can shed light on designing generalizable fraud detection algorithms in the future. 

\begin{table}[]
    \centering
        \caption{Cross-{type} validation between two driver's license  datasets. }
    \begin{tabularx}{\textwidth}{cc *{2}{>{\centering\arraybackslash}X}}
    \toprule
     & & \multicolumn{2}{c}{Test Data {Type}}\\
    \cmidrule(lr){3-4}
     Training Data {Type} & Task & WV {Driver's License} & AZ {Driver's License}\\
    \midrule
     \multirow{4}{*}{WV {Driver's License}} & Face Morphing & 98.65 & 50.55 \\
     & Portrait substitution & 88.95 & 71.20 \\
     & Mixed Fraud Face & 89.30 & 71.15 \\
     & Text Replacement & 100.00 & 60.92 \\
     \midrule
     \multirow{4}{*}{AZ {Driver's License}} & Face Morphing & 54.10 & 98.10 \\
     & Portrait substitution & 80.05 & 88,75 \\
     & Mixed Fraud Face & 71.10 & 88.80 \\
     & Text Replacement & 50.00 & 100.00 \\
    \bottomrule
    \end{tabularx}
    \label{tab:cross_domain_val}
\end{table}

\eat{
\begin{table}[]
    \centering
        \caption{Cross-domain validation between two driver license datasets. }
    \begin{tabularx}{\textwidth}{cc *{2}{>{\centering\arraybackslash}X}}
    \toprule
     & & \multicolumn{2}{c}{Test Data Domain}\\
    \cmidrule(lr){3-4}
     Training Data Domain & Task & WV Driver License & AZ Driver License \\
    \midrule
     \multirow{4}{*}{AZ Driver License} & Face Morphing & 54.10 & 98.10 \\
     & Portrait substitution & 80.05 & 88,75 \\
     & Mixed Fraud Face & 71.10 & 88.80 \\
     & Text Replacement & 50.00 & 100.00 \\
    \bottomrule
    \end{tabularx}
    \label{tab:cross_domain_val}
\end{table}
}

\begin{table}[]
    \centering
        \caption{Cross-domain validation between two ID card datasets. }
    \begin{tabularx}{\textwidth}{cc *{2}{>{\centering\arraybackslash}X}}
    \toprule
     & & \multicolumn{2}{c}{Test Data Domain}\\
    \cmidrule(lr){3-4}
     Training Data Domain & Task & Spain ID Card & Finland ID Card \\
    \midrule
     \multirow{4}{*}{Finland ID Card} & Face Morphing & 78.80 & 100.00 \\
     & Portrait substitution & 55.25 & 100.00 \\
     & Mixed Fraud Face & 49.00 & 100.00 \\
     & Text Replacement & 50.01 & 100.00 \\
    \bottomrule
    \end{tabularx}
    \vspace{-10pt}
    \label{tab:cross_domain_val-id}
\end{table}

\begin{table}[]
    \centering
        \caption{Cross-domain validation between two passport datasets. }
    \begin{tabularx}{\textwidth}{cc *{2}{>{\centering\arraybackslash}X}}
    \toprule
     & & \multicolumn{2}{c}{Test Data Domain}\\
    \cmidrule(lr){3-4}
     Training Data Domain & Task & Albania Passport & Greece Passport \\
    \midrule
     \multirow{4}{*}{Greece Passport} & Face Morphing & 45.15 & 100.00 \\
     & Portrait substitution & 55.55 & 100.00 \\
     & Mixed Fraud Face & 54.65 & 100.00 \\
     & Text Replacement & 35.56 & 100.00 \\
    \bottomrule
    \end{tabularx}
     \vspace{-10pt}
    \label{tab:cross_domain_val-passport}
\end{table}

\begin{table}[]
    \centering
        \caption{Accuracy of models trained on AZ Driver's License, tested on Spain ID Card and Albania Passport. }
    \begin{tabularx}{\textwidth}{cc *{3}{>{\centering\arraybackslash}X}}
    \toprule
     & & \multicolumn{3}{c}{Test Data Domain}\\
    \cmidrule(lr){3-5}
     Training Data Domain & Task & Spain ID Card & Albania Passport & AZ Driver License \\
    \midrule
     \multirow{4}{*}{AZ Driver License} & Face Morphing & 51.10 & 54.30 & 98.10 \\
     & Portrait substitution & 64.85 & 64.15 & 88.75 \\
     & Mixed Fraud Face & 60.35 & 40.20 & 88.80 \\
     & Text Replacement & 57.94 & 58.34 & 100.00 \\
    \bottomrule
    \end{tabularx}
     \vspace{-10pt}
    \label{tab:cross_domain_val-az-file-types}
\end{table}

\begin{table}[]
    \centering
        \caption{Accuracy of models trained on Finland ID Card, tested on Albania Passport and AZ Driver's License.  }
    \begin{tabularx}{\textwidth}{cc *{3}{>{\centering\arraybackslash}X}}
    \toprule
     & & \multicolumn{3}{c}{Test Data Domain}\\
    \cmidrule(lr){3-5}
     Training Data Domain & Task & Albania Passport & AZ Driver License & Finland ID Card\\
    \midrule
     \multirow{4}{*}{Finland ID Card} & Face Morphing & 87.95 & 57.20 & 100.00 \\
     & Portrait substitution & 64.90 & 52.40 & 100.00 \\
     & Mixed Fraud Face & 13.35 & 49.80 & 100.00 \\
     & Text Replacement & 55.52 & 50.00 & 100.00 \\
    \bottomrule
    \end{tabularx}
     \vspace{-10pt}
    \label{tab:cross_domain_val-finland-file-types}
\end{table}

\begin{table}[]
    \centering
        \caption{Accuracy of models trained on Greece Passport, tested on Spain ID Card and AZ Driver's License. }
    \begin{tabularx}{\textwidth}{cc *{3}{>{\centering\arraybackslash}X}}
    \toprule
     & & \multicolumn{3}{c}{Test Data Domain}\\
    \cmidrule(lr){3-5}
     Training Data Domain & Task & Spain ID Card & AZ Driver License & Greece Passport \\
    \midrule
     \multirow{4}{*}{Greece Passport} & Face Morphing & 53.55 & 62.00 & 100.00 \\
     & Portrait substitution & 58.50 & 61.80 & 100.00 \\
     & Mixed Fraud Face & 53.20 & 58.55 & 100.00 \\
     & Text Replacement & 39.57 & 50.00 & 100.00 \\
    \bottomrule
    \end{tabularx}
     \vspace{-10pt}
    \label{tab:cross_domain_val-Greece-Passport}
\end{table}

As mentioned, the cross-type analysis challenges were also observed in the face morphing task when applied to identity documents from different types. 
Our novel IDNet benchmark has IDs belonging to $20$ different types. Although the photos of these IDs are from the same generated dataset~\cite{generated.photos}, each ID type has a unique background, as well as different requirements and formats on the ID photo. For example, the portrait photo on the Arizona Driver's License is transparent and monochrome. In contrast, the portrait photo on the West Virginia Driver's License and Greece Passport is opaque and colored. 
In addition, the facial image on the Finland ID is cropped, opaque, and monochrome. 
We studied how face morphing detection methods transfer between different ID types and the domain transfer capabilities of PW-MAD, Inception, and MixFaceNet-S on three different types of IDs with different facial image requirements. 
We trained the models on each ID type and evaluated the models on all other ID types. The results are shown in Table~\ref{tab:face_morphing_cross}. We observed that all evaluated face morphing detection models could not transfer well to other domains (i.e., document types). The performance degradation suggests a research gap for face morphing detection with good transferability. It also showed that the IDNet dataset has the ability to challenge and motivate the cross-type face morphing detection research.

\begin{table}[!htbp]

\centering
\caption{Cross-domain evaluation between different types of ID for face morphing detection using different algorithms.}
\scalebox{1}{
\begin{tabularx}{\textwidth}{ccc|c|c}

\toprule
\multirow{2}{*}{Training Datasets} & \multirow{2}{*}{Models} & \multicolumn{3}{c}{Test Datasets} \\ \cmidrule{3-5} 
                                   &                         & AZ Driver's License & Greece Passport & Finland ID \\ \midrule
\multirow{3}{*}{AZ Driver's License} & PW-MAD                  & 97.8                & 53.4            & 58.5       \\
                                     & Inception               & 97.6                & 54.1            & 59.8       \\
                                     & MixFaceNet-S            & 98.1                & 53.5            & 61.7       \\ \midrule
\multirow{3}{*}{Greece Passport}     & PW-MAD                  & 61.6                & 100.0           & 59.7       \\
                                     & Inception               & 62.1                & 99.9            & 60.2       \\
                                     & MixFaceNet-S            & 62.0                & 100.0           & 60.5       \\ \midrule
\multirow{3}{*}{Finland ID}          & PW-MAD                  & 56.9                & 54.7            & 99.9       \\
                                     & Inception               & 58.4                & 55.9            & 99.7       \\
                                     & MixFaceNet-S            & 57.2                & 55.4            & 100.0      \\ \bottomrule
\end{tabularx}
}
\label{tab:face_morphing_cross}
\end{table}
\subsection{LLM-based Schema Alignment and Unification}

As illustrated in Fig.~\ref{fig:templates}, different types of identity documents are distinct from each other in metadata fields and the usage of security features. This leads to several challenges~\cite{wang2015schema}\cite{sharma2023automatic}\cite{shi2018method}\cite{wang2020integration}\cite{wang2020survive}: \textbf{(1) Data management challenges.} Many organizations need to manage copies of identity documents of employees, customers, and residents, who are from different states and even different countries. 
Without a unified schema, we can not easily perform analytics queries over diverse identity documents, e.g., to find all documents that use a security feature that was recently deprecated (because of emerging attack techniques); and to find anomaly identity documents that have duplicate document identifier numbers, which is an indication of fraud~\cite{onfido}. \textbf{(2) Model training challenges.} Machine learning techniques widely used to analyze documents and detect frauds, require abundant features to achieve acceptable accuracy. Extracting features from diverse identity documents that have heterogeneous schemas is also a tedious process that requires a lot of manual data integration effort. \textbf{(3) Data exchange and interoperability challenges.} Organization needs a unified schema to share data with each other. 

Limited schema alignment and metadata support were provided in today’s identity document management {designs}. There exist tremendous international efforts to standardize the design and production of identity documents {for specific document types such as a passport.}. For example, ICAO 9303 (ISO/IEC 7501)~\cite{icao} was published to standardize the machine-readable passport, visas, and official travel documents. However, these standards mostly focus on the design and implementation of the identity documents, such as physical characteristics, languages/characters/typeface/font, representations of fields such as issuing state, organization, nationality, place-of-birth, dates, and the constraints of the machine-readable zone. Other standards~\cite{aamva1} defined security features, e.g., guilloche patterns, chips, holographic overlays, watermarks, ghost images, and micro-printing. However, none of these standards provide a unified electronic and structured schema to describe all meta information about an identity document. As a result, it is challenging to align and integrate different document types (e.g., the field of first name in an Arizona Driver’s License maps to the field of Primer Apellido in a Spanish ID card). Lacking such cross-type schema alignment and integration capabilities leads to difficulties in cross-type queries. For example, when a security feature becomes {“insecure”} due to an “attack”, it is hard to query which of the thousands of types of identity documents contain this attacked security feature and should be alerted accordingly. Lacking cross-type data integration will also result in difficulty for cross-type analysis.

 Our benchmark contains twenty different types of identity documents, and it can be used for testing and comparing various schema designs and schema alignment and transformation tools. To provide a baseline, we developed a novel methodology that leverages the amazing complex reasoning and coding capabilities of the large language model (LLM), such as ChatGPT, to automatically transform each identity document to conform to a standard schema.  
 
 First, we leverage the 
well-known National Information Exchange Model (NIEM)~\cite{domain2017national} to compose an exchangeable and extensible standard schema in JSON format. The NIEM has defined core schemas and vocabulary for entities in many domains. These core schemas provide interoperability across many systems,  while participants have the flexibility to handle local data needs by creating extensions of the core schemas~\cite{mork2009galaxy}. However, designing a data model following the NIEM standard is tedious work, as the designer needs to browse hundreds to thousands of elements to select the relevant ones. To address this issue, we first use the NIEM schema subset tool (SSGT)~\cite{ssgt} to obtain a rough subset of schema elements in XML file. Then, we provide the identity documents and the XML file to {ChatGP- 4.0-turbo, which achieved better accuracy than ChatGPT-3.5-turbo}, and generate a high-quality schema that not only conforms to the NIEM standard, but is also extended with indicators for security features. The prompt template is ''\textit{You are a driver's license examiner. You are manually extracting information from a driver's license.
Please generate a JSON representation that aligns with the following NIEM schemas \{Schema Subsets Generated by SSGT\}. 
In addition to the elements listed above, you need to extend the NIEM schema to include security features such as \{A List of security features obtained from published standards~\cite{aamva1}\cite{anti-counterfeit}\}, and use "true" or "false" to indicate if they exist in the image.
Make sure to keep the schema element names unchanged.}". 
Second, we leverage the {ChatGPT-4.0-turbo} APIs to automatically convert any incoming identity documents to conform to the JSON representation generated in the first step using a simple prompt of ''\textit{Please convert the attached Driver's license into a JSON file that conforms to the schema of the attached example. Please extend the schema if needed.}". Among $400$ testing cases randomly sampled from our benchmark dataset, this methodology achieved $94.6\%$ accuracy. The errors mainly come from three aspects: (1) OCR errors (account for $17\%$ of the errors): e.g., 5'-05'' is recognized as 5'-06''; (2) Ignoring existing field error (account for $22\%$): e.g. ChatGPT4 ignored the "RESTRICTIONS" field on some Arizona driver's licenses and the "ENDORSEMENTS" field on some Washington DC driver's licenses; (3) Guessing nonexistent field error (account for $44\%$): e.g., the North Carolina and Pennsylvania driver's licenses do not have weight information, however, ChatGPT4 would guess this information and fill it in the output; and (4) Reasoning errors (account for $17\%$ of the errors), such as considering the index of a field as the value in the field and ignoring the first name while the family name contains multiple names.

{\color{black}

\subsection{ID replacement in user-defined backgrounds}

\begin{figure}[h]
  \centering
  \begin{tabular}{ccccc}
    \raisebox{0.8\height}{\includegraphics[width=0.18\textwidth]{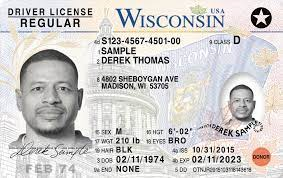}} & \hspace{0.3em}
    \includegraphics[width=0.18\textwidth]{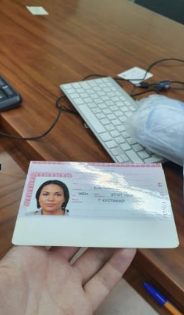} & \hspace{0.3em}
    \includegraphics[width=0.18\textwidth]{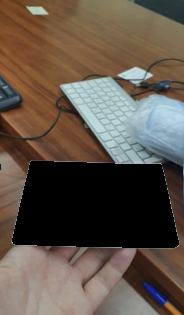} & \hspace{0.3em}
    \includegraphics[width=0.18\textwidth]{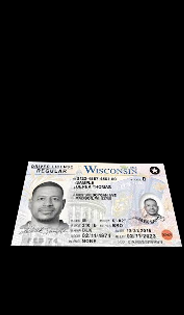} & \hspace{0.3em}
    \includegraphics[width=0.18\textwidth]{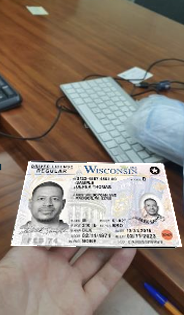} \\
    \small{(a) ID} & \small{(b) Background} & \small{(c) Masked BG} & \small{(d) Trans. ID} & \small{(e) Final Image} \\
  \end{tabular}
  \caption{ID replacement process examples.}
  \label{fig:id_replacement}
\end{figure}

One use case of the IDNet dataset is to generate mobile documents (i.e. camera captured pictures and videos that contain ID documents) with user-specified parameters, such as camera model, indoor/outdoor environment, background objects, and lighting condition. In this section, we present a simple technique to create such a dataset by replacing an ID document in an existing picture with an identity document from IDNet as shown in Figure \ref{fig:id_replacement}(e). This necessitates ensuring that the inserted ID card appears natural and consistent within the context of the original image, which involves addressing several technical challenges: accurate detection and localization of the existing ID card, precise segmentation to isolate the ID card from the background, alignment of the new ID card to match the perspective of the original, and blending to ensure a smooth transition between the new ID card and the background.
To meet these challenges, we employ a series of advanced models and image processing techniques. Grounding DINO \cite{liu2023grounding}, a state-of-the-art model that combines object detection and language grounding, is employed to accurately identify and localize the ID card within the background image. Following this, in Figure \ref{fig:id_replacement}(c) the Segment Anything Model (SAM) \cite{kirillov2023segment} is employed to segment the detected ID card from the background. SAM uses a promptable interface to identify and provide a precise mask of the ID card, facilitating accurate alignment of the new ID image.
In Figure \ref{fig:id_replacement}(d), a perspective transformation is then applied to align the new ID image with the segmented area of the original ID card. This transformation adjusts the new ID image to fit naturally within the segmented region, ensuring correct perspective, alignment and proportion. Finally, an image blending process using Laplacian pyramids is used to merge the transformed ID image with the background image seamlessly. 
}

\section{Conclusions}

In this study, we introduce IDNet, a vast and comprehensive benchmark dataset comprising $837,060$ synthetic identity cards across $20$ distinct {types}, aimed at facilitating research in identity document analysis and privacy-enhanced fraud detection. Utilizing recent breakthroughs in artificial intelligence and large language models (LLMs), we developed a cost-effective methodology for generating representative document templates and identities. Additionally, we meticulously crafted a set of representative fraud patterns—encompassing face morphing, portrait substitution, text field alterations, and combinations thereof—that are intricately designed to intersect with personal identifier information. This intersection presents significant challenges for the development of privacy-preserving fraud detection mechanisms, as evidenced by our detailed evaluations.

Furthermore, the IDNet dataset serves as a critical tool for benchmarking various privacy-preserving fraud detection algorithms, exploring the complexities of cross-type fraud detection. {We also identified the importance of providing a unified schema for diverse identity document types.}

To the best of our knowledge, IDNet stands as the most extensive publicly accessible synthetic dataset for identity document benchmarks to date. Looking ahead, we plan to (1) expand IDNet by incorporating additional fraud patterns and extending its application to a broader range of (privacy-preserving) fraud detection and document analysis scenarios; (2) extend the AI-assisted identity document generation pipeline to mobile scenarios, {so that generated documents simulate images captured using different mobile devices and lighting conditions}; (3) investigate new transfer learning or federated learning methods to facilitate cross-type knowledge transferring and build universal models on identity documents integrated using unified schemas; and (4) {continue to explore schema unification and standardization for various types of identity documents including metadata and their security features.}



\section{Acknowledgements}
This material is based upon work supported by the U.S. Department of Homeland Security under Grant Award Number 17STQAC00001-07-00.

\section{Disclaimer}
The views and conclusions contained in this document are those of the authors and should not be interpreted as necessarily representing the official policies, either expressed or implied, of the U.S. Department of Homeland Security.

%
%
%

%
\appendix
\section{Entropy of Generated Text Fields}
\label{sec:entropy}
In Section \ref{Metadata Diversity}, Slovak ID documents are used to show the entropies of the generated text fields for MIDV-2020 and IDNet. The rest of the nine countries have similar entropies. Figure \ref{fig:entropy_barbplot} shows the full results for eight other European countries. Russian ID documents are excluded because there is not expiry information on the document.

\begin{figure}[h]
\begin{center}
  \includegraphics[width=\linewidth]{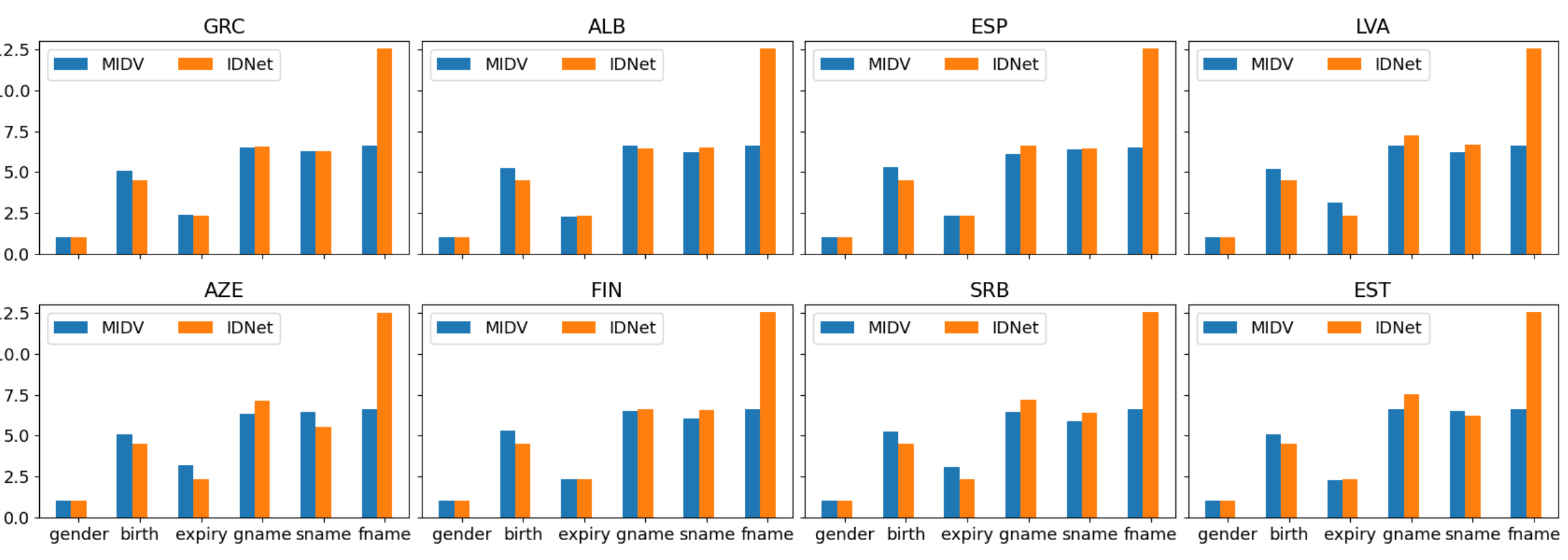}
\end{center}
\vspace{-2mm}
\caption{Comparison of generated text field entropy between MIDV and IDNet for ID documents of eight European countries. "birth" stands for the birth year, "expiry" stands for the expiry year, "gname" stands for the given name, "sname" stands for the surname, "fname" stands for the full name. 
}
\label{fig:entropy_barbplot}
\end{figure}

\bibliographystyle{splncs04}
\bibliography{refs}

\begin{thebibliography}{10}
\providecommand{\url}[1]{\texttt{#1}}
\providecommand{\urlprefix}{URL }
\providecommand{\doi}[1]{https://doi.org/#1}

\bibitem{ethnicity-groups}
About the topic of race, \url{https://www.census.gov/topics/population/race/a bout.html}

\bibitem{generated.photos}
Academic dataset by generated photos. \url{https://generated.photos/datasets}

\bibitem{age-verification}
Age verification for social media. \url{https://www.yoti.com/social-media/}

\bibitem{aws-pricing}
Aws pricing, \url{https://aws.amazon.com/ec2/pricing/on-demand/}

\bibitem{masking2}
Dod 5220.22-m – the secure wiping standard to get rid of data. \url{https://www.bitraser.com/blog/dod-wiping-the-secure-wiping-standard-to-get-rid-of-data/}

\bibitem{morphing-biometrics}
Face morphing threat to biometric identity credentials’ trustworthiness a growing problem, \url{https://www.biometricupdate.com/202010/face-morphing-threat-to-biometric-identity-credentials-trustworthiness-a-growing-problem}

\bibitem{FinCEN}
Fincen issues analysis of identity-related suspicious activities, \url{https://www.fincen.gov/news/news-releases/fincen-issues-analysis-identity-related-suspicious-activity}

\bibitem{frgc}
Frgc-morphs dataset. \url{https://www.nist.gov/programs-projects/face-recognition-grand-challenge-frgc}

\bibitem{frll}
Frll-morphs dataset. \url{https://figshare.com/articles/dataset/Face_Research_Lab_London_Set/5047666}

\bibitem{icao}
Icao 9303: Machine readable passports, \url{https://www.icao.int/publications/pages/publication.aspx?docnum=9303}

\bibitem{ssgt}
Schema subset generation tool (ssgt). \url{https://niem.github.io/reference/tools/ssgt/}

\bibitem{aamva1}
Aamva dl and id card design standard (2020). \url{https://www.aamva.org/assets/best-practices,-guides,-standards,-manuals,-whitepapers/aamva-dl-id-card-design-standard-(2020)} (2020)

\bibitem{anti-counterfeit}
Anti-counterfeiting technology guide. \url{https://euipo.europa.eu/tunnel-web/secure/webdav/guest/document\_library/observatory/documents/reports/2021 \_Anti\_Counterfeiting\_Technology\_Guide/2021\_Anti\_Counterfeiting\_Technology\_Guide \_en.pdf} (2021)

\bibitem{portrait-substitution1}
Document liveness detection and its role in preventing identity fraud. \url{https://www.idrnd.ai/document-fraud-in-digital-onboarding/} (2023)

\bibitem{onfido}
Identity fraud report 2024. \url{https://onfido.com/landing/identity-fraud-report/} (2023)

\bibitem{abedjan2015profiling}
Abedjan, Z., Golab, L., Naumann, F.: Profiling relational data: a survey. The VLDB Journal  \textbf{24},  557--581 (2015)

\bibitem{al2023guilloche}
Al-Ghadi, M., Ming, Z., Gomez-Kr{\"a}mer, P., Burie, J.C., Coustaty, M., Sidere, N.: Guilloche detection for id authentication: A dataset and baselines. In: IEEE 25th International Workshop on Multimedia Signal Processing (2023)

\bibitem{alghofaili2020financial}
Alghofaili, Y., Albattah, A., Rassam, M.A.: A financial fraud detection model based on lstm deep learning technique. Journal of Applied Security Research  \textbf{15}(4),  498--516 (2020)

\bibitem{arlazarov2019midv}
Arlazarov, V.V., Bulatov, K.B., Chernov, T.S., Arlazarov, V.L.: Midv-500: a dataset for identity document analysis and recognition on mobile devices in video stream. Computer Optics  \textbf{43}(5),  818--824 (2019)

\bibitem{bernabe2020aries}
Bernabe, J.B., David, M., Moreno, R.T., Cordero, J.P., Bahloul, S., Skarmeta, A.: Aries: Evaluation of a reliable and privacy-preserving european identity management framework. Future Generation Computer Systems  \textbf{102},  409--425 (2020)

\bibitem{boned2024synthetic}
Boned, C., Talarmain, M., Ghanmi, N., Chiron, G., Biswas, S., Awal, A.M., Terrades, O.R.: Synthetic dataset of id and travel document. arXiv preprint arXiv:2401.01858  (2024)

\bibitem{boutros2021mixfacenets}
Boutros, F., Damer, N., Fang, M., Kirchbuchner, F., Kuijper, A.: Mixfacenets: Extremely efficient face recognition networks. In: 2021 IEEE International Joint Conference on Biometrics (IJCB). pp.~1--8. IEEE (2021)

\bibitem{bulatov2020midv}
Bulatov, K., Matalov, D., Arlazarov, V.V.: Midv-2019: challenges of the modern mobile-based document ocr. In: Twelfth International Conference on Machine Vision (ICMV 2019). vol. 11433, pp. 717--722. SPIE (2020)

\bibitem{bulatovich2022midv}
Bulatovich, B.K., Vladimirovna, E.E., Vyacheslavovich, T.D., Sergeevna, S.N., Sergeevna, C.Y., Zuheng, M., Jean-Christophe, B., Muzzamil, L.M.: Midv-2020: a comprehensive benchmark dataset for identity document analysis. Computer Optics  \textbf{46}(2),  252--270 (2022)

\bibitem{chazalon2017smartdoc}
Chazalon, J., Gomez-Kr{\"a}mer, P., Burie, J.C., Coustaty, M., Eskenazi, S., Luqman, M., Nayef, N., Rusi{\~n}ol, M., Sidere, N., Ogier, J.M.: Smartdoc 2017 video capture: Mobile document acquisition in video mode. In: 2017 14th IAPR International Conference on Document Analysis and Recognition (ICDAR). vol.~4, pp. 11--16. IEEE (2017)

\bibitem{chopra2023cowrangler}
Chopra, B., Fariha, A., Gulwani, S., Henley, A.Z., Perelman, D., Raza, M., Shi, S., Simmons, D., Tiwari, A.: Cowrangler: Recommender system for data-wrangling scripts. In: Companion of the 2023 International Conference on Management of Data. pp. 147--150 (2023)

\bibitem{damer2022privacy}
Damer, N., L{\'o}pez, C.A.F., Fang, M., Spiller, N., Pham, M.V., Boutros, F.: Privacy-friendly synthetic data for the development of face morphing attack detectors. In: Proceedings of the IEEE/CVF Conference on Computer Vision and Pattern Recognition. pp. 1606--1617 (2022)

\bibitem{damer2021pw}
Damer, N., Spiller, N., Fang, M., Boutros, F., Kirchbuchner, F., Kuijper, A.: Pw-mad: Pixel-wise supervision for generalized face morphing attack detection. In: Advances in Visual Computing: 16th International Symposium, ISVC 2021, Virtual Event, October 4-6, 2021, Proceedings, Part I. pp. 291--304. Springer (2021)

\bibitem{domain2017national}
Domain, M.O., Plan, S.: National information exchange model (niem)  (2017)

\bibitem{el2020practical}
El~Emam, K., Mosquera, L., Hoptroff, R.: Practical synthetic data generation: balancing privacy and the broad availability of data. O'Reilly Media (2020)

\bibitem{fang2022unsupervised}
Fang, M., Boutros, F., Damer, N.: Unsupervised face morphing attack detection via self-paced anomaly detection. In: 2022 IEEE International Joint Conference on Biometrics (IJCB). pp. 1--11. IEEE (2022)

\bibitem{frazier2018tutorial}
Frazier, P.I.: A tutorial on bayesian optimization. arXiv preprint arXiv:1807.02811  (2018)

\bibitem{fritz2022financial}
Fritz-Morgenthal, S., Hein, B., Papenbrock, J.: Financial risk management and explainable, trustworthy, responsible ai. Frontiers in artificial intelligence  \textbf{5},  779799 (2022)

\bibitem{ho2020denoising}
Ho, J., Jain, A., Abbeel, P.: Denoising diffusion probabilistic models. Advances in neural information processing systems  \textbf{33},  6840--6851 (2020)

\bibitem{hore2010image}
Hore, A., Ziou, D.: Image quality metrics: Psnr vs. ssim. In: 2010 20th international conference on pattern recognition. pp. 2366--2369. IEEE (2010)

\bibitem{jordon2022synthetic}
Jordon, J., Szpruch, L., Houssiau, F., Bottarelli, M., Cherubin, G., Maple, C., Cohen, S.N., Weller, A.: Synthetic data--what, why and how? arXiv preprint arXiv:2205.03257  (2022)

\bibitem{kamuangu2024review}
Kamuangu, P.: A review on financial fraud detection using ai and machine learning. Journal of Economics, Finance and Accounting Studies  \textbf{6}(1),  67--77 (2024)

\bibitem{kirillov2023segment}
Kirillov, A., Mintun, E., Ravi, N., Mao, H., Rolland, C., Gustafson, L., Xiao, T., Whitehead, S., Berg, A.C., Lo, W.Y., et~al.: Segment anything. In: Proceedings of the IEEE/CVF International Conference on Computer Vision. pp. 4015--4026 (2023)

\bibitem{korshunov2013using}
Korshunov, P., Ebrahimi, T.: Using face morphing to protect privacy. In: 2013 10th IEEE International Conference on Advanced Video and Signal Based Surveillance. pp. 208--213. IEEE (2013)

\bibitem{lecuyer2019certified}
Lecuyer, M., Atlidakis, V., Geambasu, R., Hsu, D., Jana, S.: Certified robustness to adversarial examples with differential privacy. In: 2019 IEEE symposium on security and privacy (SP). pp. 656--672. IEEE (2019)

\bibitem{liu2023grounding}
Liu, S., Zeng, Z., Ren, T., Li, F., Zhang, H., Yang, J., Li, C., Yang, J., Su, H., Zhu, J., et~al.: Grounding dino: Marrying dino with grounded pre-training for open-set object detection. arXiv preprint arXiv:2303.05499  (2023)

\bibitem{mercer1998document}
Mercer, J.W.: Document fraud deterrent strategies: four case studies. In: Optical Security and Counterfeit Deterrence Techniques II. vol.~3314, pp. 39--51. SPIE (1998)

\bibitem{mork2009galaxy}
Mork, P., Seligman, L., Rosenthal, A., Morse, M., Wolf, C., Hoyt, J., Smith, K.: Galaxy: Encouraging data sharing among sources with schema variants. In: 2009 IEEE 25th International Conference on Data Engineering. pp. 1551--1554. IEEE (2009)

\bibitem{ngan2022face}
Ngan, M., Grother, P., Hanaoka, K., Kuo, J.: face recognition vendor test (frvt): Morph-performance of automated face morph detection. National Institute of Technology (NIST)  (2022)

\bibitem{ngan2020face}
Ngan, M., Ngan, M., Grother, P., Hanaoka, K., Kuo, J.: Face recognition vendor test (frvt) part 4: Morph-performance of automated face morph detection  (2020)

\bibitem{ngoc2018saliency}
Ngoc, M.{\^O}.V., Fabrizio, J., G{\'e}raud, T.: Saliency-based detection of identy documents captured by smartphones. In: 2018 13th IAPR international workshop on document analysis systems (DAS). pp. 387--392. IEEE (2018)

\bibitem{ramachandra2019detecting}
Ramachandra, R., Venkatesh, S., Raja, K., Busch, C.: Detecting face morphing attacks with collaborative representation of steerable features. In: Proceedings of 3rd International Conference on Computer Vision and Image Processing: CVIP 2018, Volume 1. pp. 255--265. Springer (2019)

\bibitem{ren2018learning}
Ren, Z., Lee, Y.J., Ryoo, M.S.: Learning to anonymize faces for privacy preserving action detection. In: Proceedings of the european conference on computer vision (ECCV). pp. 620--636 (2018)

\bibitem{rombach2022high}
Rombach, R., Blattmann, A., Lorenz, D., Esser, P., Ommer, B.: High-resolution image synthesis with latent diffusion models. In: Proceedings of the IEEE/CVF conference on computer vision and pattern recognition. pp. 10684--10695 (2022)

\bibitem{de2020bid}
de~S{\'a}~Soares, A., das Neves~Junior, R.B., Bezerra, B.L.D.: Bid dataset: a challenge dataset for document processing tasks. In: Anais Estendidos do XXXIII Conference on Graphics, Patterns and Images. pp. 143--146. SBC (2020)

\bibitem{sarkar2020vulnerability}
Sarkar, E., Korshunov, P., Colbois, L., Marcel, S.: Vulnerability analysis of face morphing attacks from landmarks and generative adversarial networks. arXiv preprint arXiv:2012.05344  (2020)

\bibitem{sarkar2022gan}
Sarkar, E., Korshunov, P., Colbois, L., Marcel, S.: Are gan-based morphs threatening face recognition? In: ICASSP 2022-2022 IEEE International Conference on Acoustics, Speech and Signal Processing (ICASSP). pp. 2959--2963. IEEE (2022)

\bibitem{setiadi2021psnr}
Setiadi, D.R.I.M.: Psnr vs ssim: imperceptibility quality assessment for image steganography. Multimedia Tools and Applications  \textbf{80}(6),  8423--8444 (2021)

\bibitem{sharma2023automatic}
Sharma, A., Li, X., Guan, H., Sun, G., Zhang, L., Wang, L., Wu, K., Cao, L., Zhu, E., Sim, A., et~al.: Automatic data transformation using large language model: An experimental study on building energy data. arXiv e-prints pp. arXiv--2309 (2023)

\bibitem{shi2018method}
Shi, J.W., Wang, C., Wang, L.J., Zou, J.: Method and apparatus for generating schema of non-relational database (Jun~19 2018), uS Patent 10,002,142

\bibitem{song2020denoising}
Song, J., Meng, C., Ermon, S.: Denoising diffusion implicit models. arXiv preprint arXiv:2010.02502  (2020)

\bibitem{tan2019mixconv}
Tan, M., Le, Q.V.: Mixconv: Mixed depthwise convolutional kernels. arXiv preprint arXiv:1907.09595  (2019)

\bibitem{venkatesh2021face}
Venkatesh, S., Ramachandra, R., Raja, K., Busch, C.: Face morphing attack generation and detection: A comprehensive survey. IEEE transactions on technology and society  \textbf{2}(3),  128--145 (2021)

\bibitem{wang2021research}
Wang, H., Guo, L.: Research on face recognition based on deep learning. In: 2021 3rd International Conference on Artificial Intelligence and Advanced Manufacture (AIAM). pp. 540--546. IEEE (2021)

\bibitem{wang2015schema}
Wang, L., Zhang, S., Shi, J., Jiao, L., Hassanzadeh, O., Zou, J., Wangz, C.: Schema management for document stores. Proceedings of the VLDB Endowment  \textbf{8}(9),  922--933 (2015)

\bibitem{wang2020survive}
Wang, Z., Zhou, L., Das, A., Dave, V., Jin, Z., Zou, J.: Survive the schema changes: integration of unmanaged data using deep learning. arXiv preprint arXiv:2010.07586  (2020)

\bibitem{wang2020integration}
Wang, Z., Zhou, L., Zou, J.: Integration of fast-evolving data sources using a deep learning approach. In: Software Foundations for Data Interoperability and Large Scale Graph Data Analytics: 4th International Workshop, SFDI 2020, and 2nd International Workshop, LSGDA 2020, held in Conjunction with VLDB 2020, Tokyo, Japan, September 4, 2020, Proceedings 4. pp. 172--186. Springer (2020)

\bibitem{wei2022emergent}
Wei, J., Tay, Y., Bommasani, R., Raffel, C., Zoph, B., Borgeaud, S., Yogatama, D., Bosma, M., Zhou, D., Metzler, D., et~al.: Emergent abilities of large language models. arXiv preprint arXiv:2206.07682  (2022)

\bibitem{wolberg1998image}
Wolberg, G.: Image morphing: a survey. The visual computer  \textbf{14}(8-9),  360--372 (1998)

\bibitem{yang2023diffusion}
Yang, L., Zhang, Z., Song, Y., Hong, S., Xu, R., Zhao, Y., Zhang, W., Cui, B., Yang, M.H.: Diffusion models: A comprehensive survey of methods and applications. ACM Computing Surveys  \textbf{56}(4),  1--39 (2023)

\bibitem{zhang2016joint}
Zhang, K., Zhang, Z., Li, Z., Qiao, Y.: Joint face detection and alignment using multitask cascaded convolutional networks. IEEE signal processing letters  \textbf{23}(10),  1499--1503 (2016)

\bibitem{zhou2023privacy}
Zhou, L., Yu, L., Zou, J., Min, H.: Privacy-preserving redaction of diagnosis data through source code analysis. In: Proceedings of the 35th International Conference on Scientific and Statistical Database Management. pp.~1--4 (2023)

\end{thebibliography}

\end{document}